%% file: main.tex
\documentclass{article}
\usepackage{amssymb}
\usepackage{booktabs}
\usepackage{graphicx}
\usepackage{float}
\usepackage{multirow}
\usepackage{makecell}
\usepackage{subcaption}
\usepackage{amsmath}
\usepackage{wrapfig}
\usepackage[table]{xcolor}
\usepackage{amssymb}       
\usepackage{caption}
\usepackage{adjustbox} 
\usepackage{pifont}
\newcommand{\xmark}{\ding{55}}
\newcommand{\cmark}{\ding{51}}
\usepackage[ruled,vlined]{algorithm2e}
\usepackage[final]{corl_2025} 

\newcommand{\norm}[1]{\left\lVert#1\right\rVert}

\DeclareMathOperator*{\argmin}{arg\,min}

\title{Co-Design of Soft Gripper with Neural Physics}

\author{Sha Yi\thanks{These authors contributed equally to this work.},\;
Xueqian Bai\footnotemark[1], \;
Adabhav Singh, \;
Jianglong Ye, \;
Michael T. Tolley, \;
Xiaolong Wang}

\begin{document}
\maketitle
\vspace{-1em}
\begin{center}
\vspace{-15pt}
  University of California, San Diego
\end{center}

\begin{figure}[H]
  \vspace{-5pt}
  \centering
\includegraphics[width=0.95\textwidth]{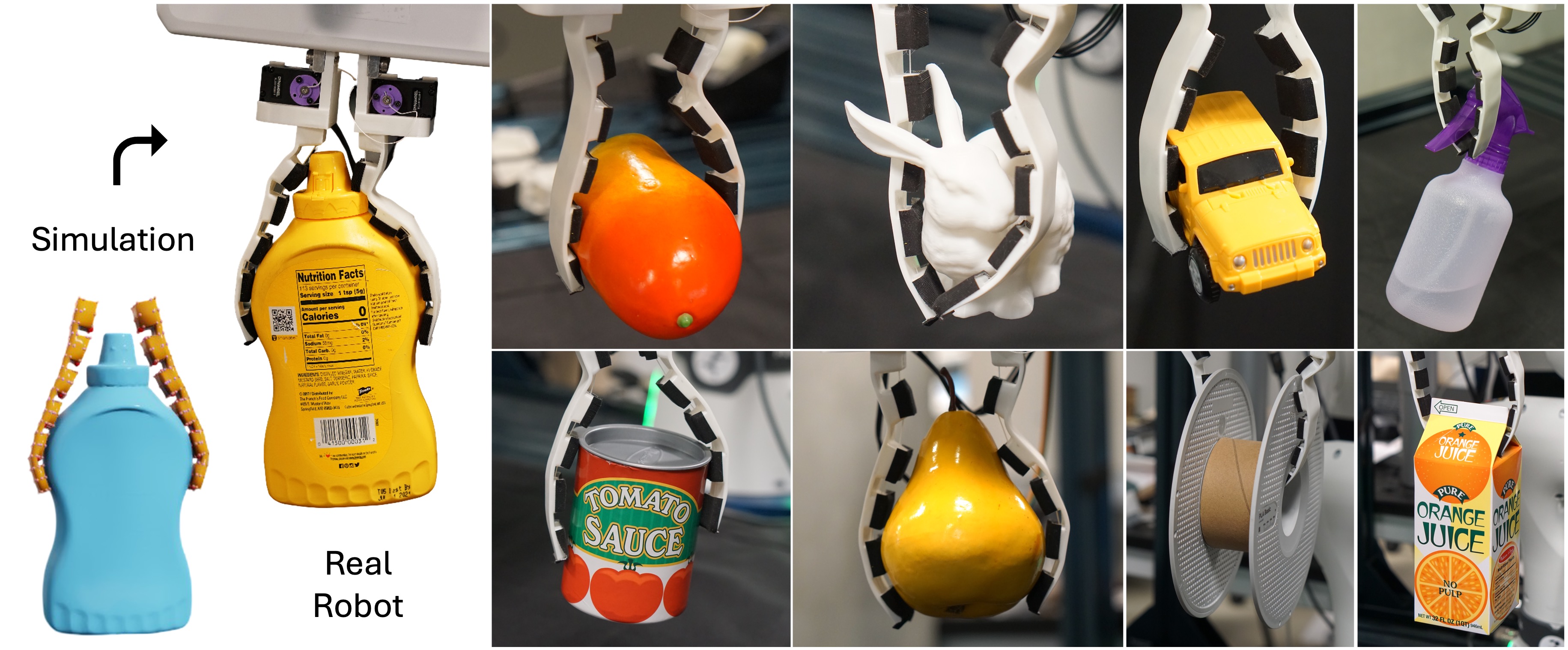}
\caption{We introduce a co-design framework that jointly optimizes the spatial stiffness distribution and grasping poses of soft grippers via a simulation-trained neural surrogate. Hardware experiments demonstrate that our optimized grippers outperform both rigid and overly compliant designs.}
\label{fig:teaser}
  \vspace{-5pt}
\end{figure}

\begin{abstract}
For robot manipulation, both the controller and end-effector design are crucial. Soft grippers are generalizable by deforming to different geometries, but designing such a gripper and finding its grasp pose remains challenging. In this paper, we propose a co-design framework that generates an optimized soft gripper's block-wise stiffness distribution and its grasping pose, using a neural physics model trained in simulation. We derived a uniform-pressure tendon model for a flexure-based soft finger, then generated a diverse dataset by randomizing both gripper pose and design parameters. A neural network is trained to approximate this forward simulation, yielding a fast, differentiable surrogate. We embed that surrogate in an end-to-end optimization loop to optimize the ideal stiffness configuration and best grasp pose. Finally, we 3D-print the optimized grippers of various stiffness by changing the structural parameters. We demonstrate that our co-designed grippers significantly outperform baseline designs in both simulation and hardware experiments. More info: \url{http://yswhynot.github.io/codesign-soft/}
\end{abstract}


\section{Introduction}
Soft robotic hands~\cite{rus2015design, shintake2018soft} offer unique compliance and adaptability, making them ideal for manipulation tasks. Unlike rigid grippers, soft structures enable safer interactions with humans, and more robust and generalizable grasps across diverse object geometries due to their inherent compliance. However, how much softness is needed? An overly compliant gripper that cannot bear any load will find most manipulation tasks challenging. We hypothesize that the design of the gripper largely depends on the tasks and what control policy we use during deployment, while in turn, the control policy can be further improved based on a better design. The joint improvement and adaptation can potentially unlock new robot morphologies and significantly improve task performance~\cite{lipson2000automatic}.

Although co‐adaptation holds great promise, the combined design–control search space is both high‐dimensional and highly nonlinear, and the two components operate on vastly different timescales - design changes occur at the time of manufacturing, whereas control updates run in real time during deployment. Previous works have explored updating designs and control in parallel~\cite{xu2021end}, as a sequential graph~\cite{chen2020hardware}, or using bi-level optimization with an outer design loop and an inner control loop~\cite{schaff2023sim, he2024morph, zhao2020robogrammar}. However, tightly coupling morphology and policy often leads to excessive interdependency, causing these methods to specialize on a single object or task and limit their broader applicability. To overcome this, we require a co‐design framework that maintains adaptability across diverse objects and behaviors while still jointly exploring design and control.

In this work, we present a data-driven co-design framework based on a neural physics learned from simulation data. We introduce a monolithic, 3D-printable soft finger using flexure joints. We propose a \textit{uniform-pressure} tendon routing method for a flexure-based soft finger, allowing the finger to deform to object surfaces. We then develop a simplified tendon model in simulation, and generate data by varying both gripper pose and the spatial stiffness parameters on a set of objects. A neural network is then trained with this forward physics mapping, enabling fast, differentiable predictions of grasp results. We show that this neural physics model is \textit{adaptable and efficient}. During optimization, we co-optimize with a control (grasping pose) as outer loop and design (stiffness) as inner loop during offline optimization with a large set of objects. During deployment, we use it as an evaluator with a fixed design to obtain grasping poses. We realize the optimized design with 3D printing based on different structural stiffness. We show that our framework significantly improves the success rate of grasping in both simulation and hardware experiments.

\section{Related Work}
Robot hands have been largely inspired by human anatomy~\cite{bircher2021complex,odhner2014compliant,gao2024sim}. Such designs aim for dexterous manipulation capabilities, but often come at the cost of high mechanical and control complexity. Nonetheless, it remains an open question how much additional hardware complexity is truly justified by the performance gains on target tasks. To address this, recent works have explored computational methods that optimize design parameters, and some that jointly optimize control policies. 

\textbf{Computational Design and Co-Design.}
Computation methods for mechanical design have leveraged gradient-based optimization ~\cite{chen2018topology, chen2020design,airfoil,dafoam} and planning methods~\cite{feshbach2024algorithmic, kodnongbua2023computational} over geometry or topology to tailor structures for target tasks. However, these approaches are often computationally expensive and are hard to generalize effectively to new objects or tasks. Differentiable simulations~\cite{warp2022, hu2019difftaichi} offer promising co-design directions by embedding both morphology and controller parameters within one framework, but require smooth, tractable gradients, and are often sensitive to initialization~\cite{xu2021end, allen2022inverse, georgievpwm}. Sampling-based frameworks trade off evaluation cost for global exploration: Bayesian optimization offers sample efficiency over a low-dimensional parameter space~\cite{calandra2016bayesian, bo}, evolutionary and CMA-ES methods handle rich discrete/continuous topologies at the expense of thousands of simulations~\cite{hansen2003reducing,kulz2024optimizing}, and RL-driven co-evolution jointly adapts design and policy in an outer–inner loop at high computational cost~\cite{he2024morph,schaff2023sim}. Generative design approaches~\cite{ha2021fit2form, xu2024dynamics} have also shown great potential for geometries.
A class of Graph Neural Network (GNN)-based simulators~\cite{meshgraph,fluid} supports effective design processes by learning representations over graph nodes and edges: GD-M~\cite{inverse} for inverse design learns design parameters for high-dimensional fluid–structure interactions using Graph Networks, but can be computationally heavy. 

\textbf{Soft Robot Fabrication and Stiffness Design.}
Soft robots encode their joints directly into the body’s geometry, allowing them to be fabricated as a single, monolithic structure without separate bulky and rigid actuators~\cite{rus2015design, piazza2019century}. Molding and casting techniques~\cite{bilodeau2015monolithic, rothemund2018soft, wei2016novel} enable monolithic soft structures, while 3D printing offers rapid iteration of custom geometries~\cite{hubbard2021fully, zhai2023desktop}. Variable-stiffness materials such as layer jamming, phase-change polymers, or tunable elastomer composites, allow a single gripper morphology to exhibit different compliance profiles under external stimuli~\cite{kim2019continuously,narang2017transforming}. Designing stiffness distributions within a soft body remains challenging: most work still relies on heuristic parameters or manual tuning.

\section{Methods}
Soft robots can be manufactured with integrated compliant joints as a single piece. We utilize a simple flexure joint in Fig.~\ref{fig:tendon_fig}, which is a thin beam connecting two thicker segment blocks~\cite{howell2013introduction, dollar2010highly}. Tendon‐driven actuation further enhances these designs by shifting the heavy motors to the base/wrist of a robot hand, allowing for dexterous, lightweight fingers. A key design variable is the distribution of material stiffness, which strongly influences contact forces and grasp performance. This section introduces a simulation‐driven co‐design framework that jointly optimizes a soft gripper’s block‐wise stiffness distribution and its grasping pose.

\subsection{Tendon Model in Simulation}\label{sec:tendon_model}

\begin{figure*}
\centering
\begin{subfigure}{0.55\textwidth}
\includegraphics[width=\textwidth]{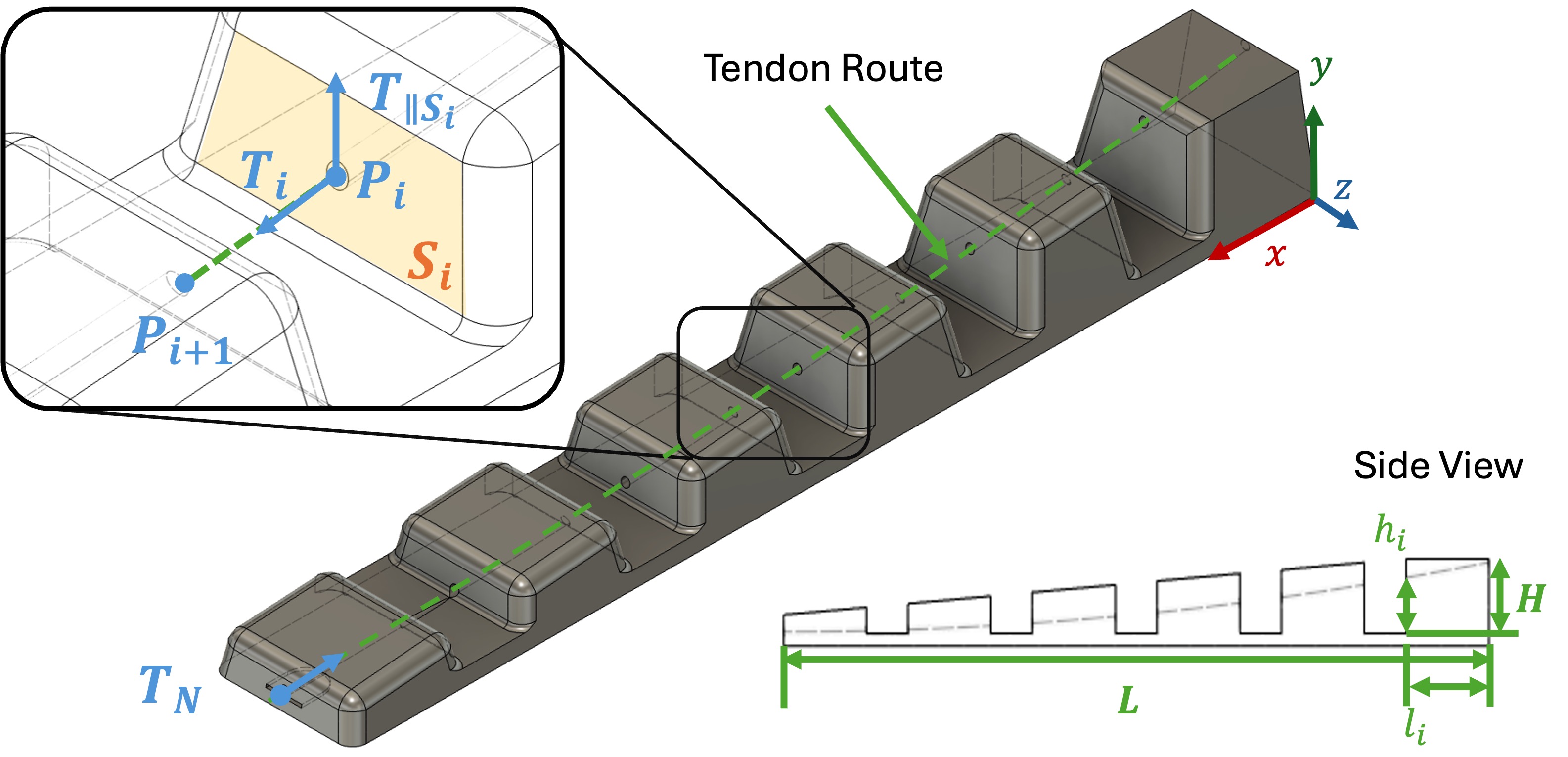}
\caption{}
\label{fig:tendon_model}
\end{subfigure}
\begin{subfigure}{0.3\textwidth}
\includegraphics[width=\textwidth]{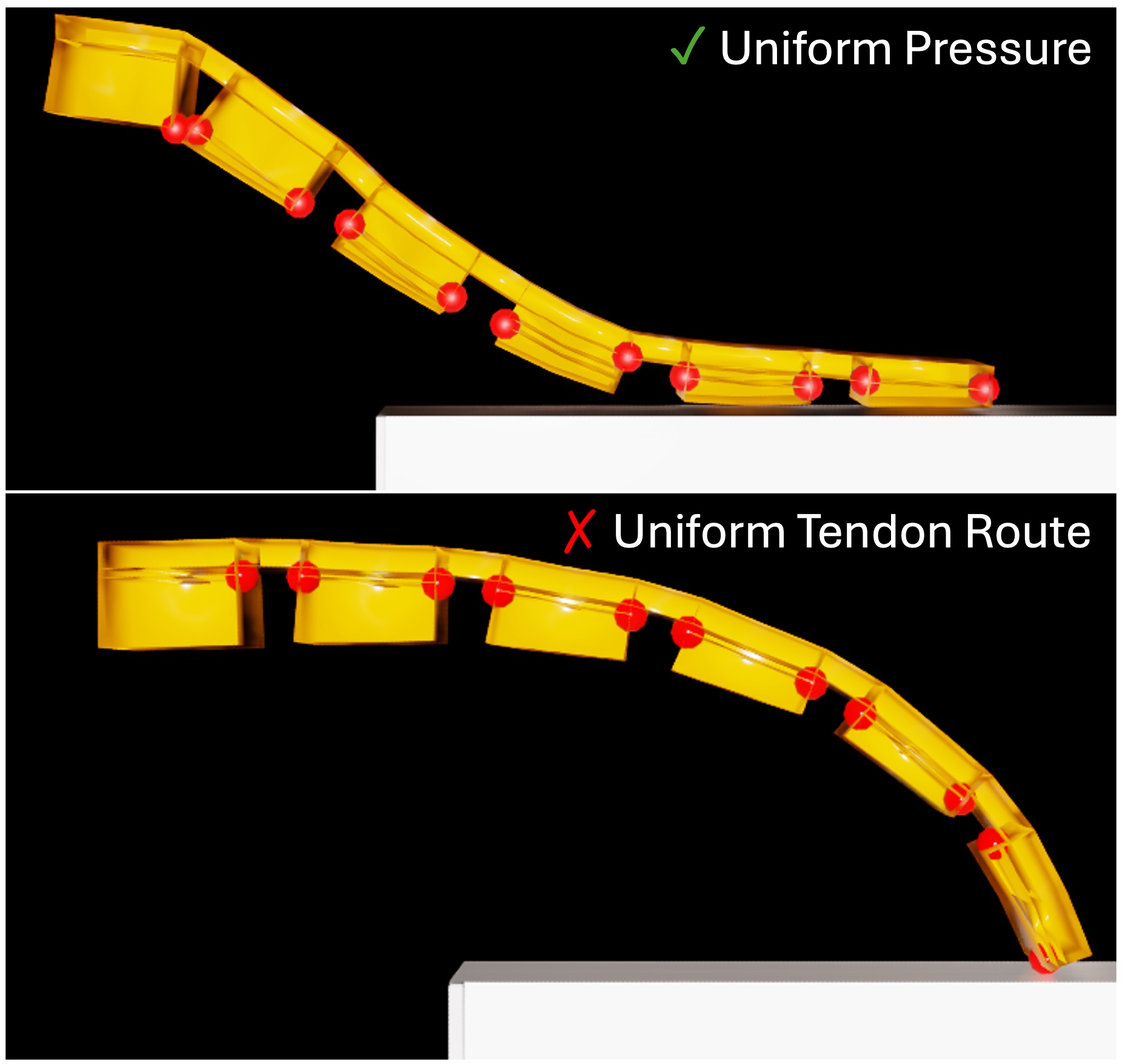}
\caption{}
\label{fig:tendon_distribution}
\end{subfigure}
\caption{(a) Tendon along a soft finger. The side view shows the dimensions of the tendon waypoints going through the soft body. (b) The influence of the tendon waypoint distribution. Top: The tendon distribution following Equation~\eqref{eq:uniform_pressure}. The finger complies with the object geometry. Bottom: The tendon waypoint is uniformly distributed along the finger. The finger will curl and only its fingertip touches the object.}
\label{fig:tendon_fig}
\vspace{-13pt}
\end{figure*}

We develop our tendon-driven soft finger simulator in Nvidia Warp~\cite{warp2022}, using the Finite-Element Method for soft bodies. We first generate a custom tetrahedral mesh of the soft body to match our gripper dimensions. Tendons are routed through this mesh via a given set of waypoints. The vertices are then modified to guarantee exactly one vertex at each tendon waypoint. An accurate, inextensible tendon model generally enforces kinematic constraints along the tendon path - constraining the sum of internal tendon displacements to match the actuator pulling distance~\cite{faure2012sofa}. However, enforcing those constraints at runtime adds significant complexity and slows down the simulation. Instead, we simplify it and only enforce that the tension force along the tendon is uniform everywhere. We therefore approximate the tendon by applying that uniform tension as concentrated forces at the mesh’s surface points in Figure~\ref{fig:tendon_model}. This ``point-force" approximation preserves the mechanical effect of an inextensible tendon while allowing us to maintain a fast, stable forward simulation.

As shown in Figure~\ref{fig:tendon_model}, we denote the tension force at waypoint $i$ to be $T_i$, where $i = 0, \cdots, N$. The waypoint at the finger tip defined as $i=N$. Therefore the tension at each waypoint is $f_T = \norm{T_0} = \cdots = \norm{T_N}$. We consider the force from the tendon acting on the soft body to be the projection of force $T_i$ onto the surface $S_i$. We define the surface normal vector of $S_i$ to be $\mathbf{n}_{S_i}$, $\norm{\mathbf{n}_{S_i}} = 1$. For a surface $S_i$ with waypoint $P_i \in \mathcal R^3$ facing the positive $x$-axis, the force acting on the soft body $T_{\parallel S_i}$ is $T_i$ projected onto the tangent plane
\begin{equation}
T_{\parallel S_i} = \bigl(I - \mathbf{n}_{S_i}\,\mathbf{n}_{S_i}^{\!\top}\bigr)T_i,\ T_i = f_T \cdot \frac{P_{i+1} - P_{i}}{\|P_{i+1}-P_i\|}\, ,  \text{  for } i < N
\label{eq:Ti_vec}
\end{equation}
Since the tendon pulls direction into the fingertip, when $i = N$ we have $T_N = - f_T \mathbf{n}_{S_N}$.

\subsection{Uniform Pressure Tendon Waypoint Distribution}\label{sec:tendon_waypoint}
The routing of a tendon significantly influences the deformation of soft fingers when in contact with an object. If the tendon is routed with the same distance to the bending flexures, the pressure will be quadratically higher when the waypoints are away from the base, as shown in the bottom example in Figure~\ref{fig:tendon_distribution}. The tip of the finger will curl when contacting the object, which is undesirable because we hope to utilize the full finger instead of only the finger tip. A uniform pressure model was initially presented on a pulley-based multi-link robot finger~\cite{hirose1978development}, and was later first used in a soft robot with pneumatic actuation~\cite{glick2018soft}. We adopt this formulation on our tendon-driven soft finger. We denote the length of one finger to be $L$, and the height of the finger base to be $H$ as shown in Figure~\ref{fig:tendon_model}. Consider a tendon waypoint $i$ with distance to the base as $l_i$ and height to its flexure joint to be $h_i$. The bending moment here should be uniform and constant, we denote it as $c$:
{\footnotesize
\begin{equation}
\frac{\mathrm{d}^2 M(l_i)}{\mathrm{d} l_i ^2} = c,\, \text{with boundary conditions }
\frac{\mathrm{d} M(L)}{\mathrm{d} L} = 0, \, M(L) = 0
\end{equation}}
The boundary conditions represent that the shear force and bending moment at the finger tip are all zeros, so the finger tip will not curl when contacting objects. Thus, the bending moment at waypoint $i$ is $M(l_i) = \tfrac{c L^2}{2}(1 - \tfrac{l_i}{L})^2$ as in~\cite{hirose1978development}. For our soft flexures and tendon routing, the bending moment $M(l_i) = h_i \times T_i$ is a linear relationship. Therefore, to have the maximum torque applied onto the object while having a uniform distribution, the tenden waypoint should follow
{\footnotesize
\begin{equation}
h_i = H\left(1 - \frac{l_i}{L}\right)^2
\label{eq:uniform_pressure}
\end{equation}}
Note that the geometry above the tendon does not directly influence the force applied to the object. However, a longer distance above the tendon route will transfer more torque from the object to the tendon itself, which gives undesirable instability. Therefore, we cut the finger blocks with a plane to minimize this extra torque, resulting in a trapezoid shape. 

\subsection{System Architecture}\label{sec:system_architecture}
\begin{figure}
\centering
\includegraphics[width=0.83\linewidth]{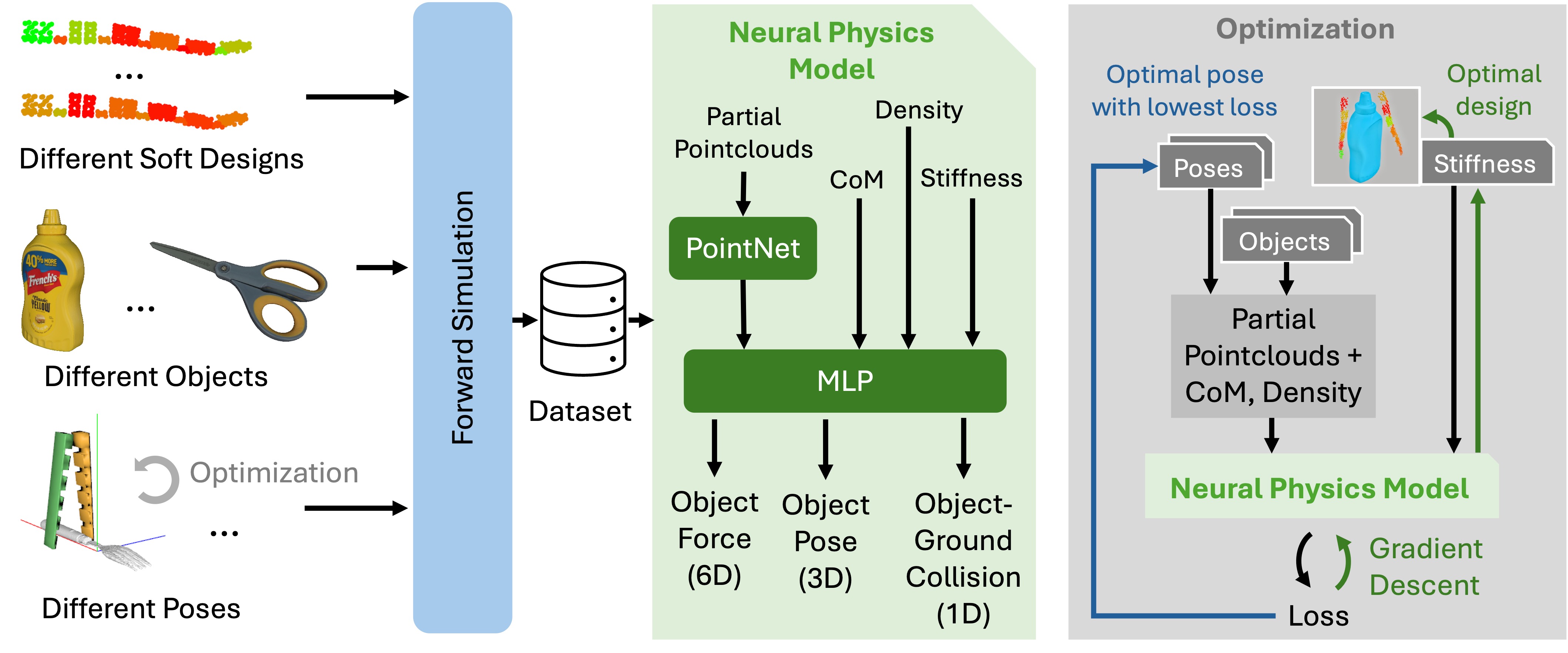}
\caption{System architecture. We first generate data by sampling different stiffness design parameters, different initial poses, and on different objects in the soft body simulation. We then use this data to train a neural network surrogate that later enables fast and accurate physical simulation. This neural physics model is then used to optimize the stiffness parameter, ultimately producing an optimal gripper design, and to evaluate sampled grasp pose candidates both during optimization and deployment.}
\label{fig:architechture}
\vspace{-10pt}
\end{figure}
As shown in Figure~\ref{fig:architechture}, we use the forward simulation we developed in Section~\ref{sec:tendon_model} to generate data. We use the YCB dataset~\cite{YCB} with daily and household objects. For each object, we sample different stiffness parameters, and initial poses based on optimization of a rigid counterpart of the soft gripper. We forward simulate this sampled setup with a fixed set of motion - pulling the tendons, then lifting the fingers up. We collect data on the ending object's body force, pose, and whether it is in contact with the ground. Neural network is shown to compile physics efficiently~\cite{son2023local}. We then train a neural network surrogate to represent the physics simulation. The input of the network is the partial point clouds, the center of mass and density of the object, and the stiffness vector. The network minimizes the L1 loss with respect to the simulation output. Compared to the slow soft body simulation, this neural network model gives a fast and reliable forward simulation, which in turn allows us to perform fast design iteration. Compared with the actual contact dynamics and gradients, the neural network is inherently smooth. We then use this model to optimize and select the optimal design parameters and gripper poses. During optimization, we use gradient descent to optimize design parameters based on the sampled pose, and use the one with the lowest loss during deployment.

\subsubsection{Data Generation and Training}\label{sec:data_generation}
We sample different block stiffnesses for the soft finger, each Young's modulus ranges from 0.7 MPa to 24 MPa. The deformation of each finger is significantly different based on the stiffness and also the geometry of object at the area of contact. Two fingers consist of 22 blocks in total, resulting in a huge design space. 
We use Anygrasp~\cite{fang2023anygrasp} to first generate a set of candidate poses from the object and ground point clouds. However, the poses from Anygrasp are very limited and do not consider the gripper length itself. Since our gripper may be longer than the Anygrasp default gripper, the direct output poses result in collisions. To provide more diverse poses that are also valid for our gripper, we augment them with optimization based on rigid counterparts of the soft gripper in simulation. The rigid counterparts are created from the soft gripper mesh, but their poses can be optimized at a much faster speed. We attach the rigid gripper with a 6D free joint to the world origin, and a prismatic joint connecting both pieces $[t_x, t_y, t_z, r_r, r_p, r_y, s_1, s_2] \in \mathcal{R}^8$. We add random normal noise to $\mathbf{t} = [t_x, t_y, t_z] + \mathcal{N}(\mu_t, \sigma_t)$ and $\mathbf{r} = [r_r, r_p, r_y] + \mathcal{N}(\mu_r, \sigma_r)$. We then optimize the gripper pose based on the signed distance function (SDF) of the finger vertices to the mesh and ground
{\footnotesize
\begin{equation}
\argmin_{t_y,s_1,s_2}
\mathcal L_{\mathrm{init}}
= w_d\sum_{i=\text{object},\text{ground}}\max\bigl(d(i),0\bigr)
+ w_p\sum_{i=\text{object},\text{ground}}\max\bigl(-d(i),0\bigr)
\end{equation}}
where $d(\cdot)$ is the SDF of the finger vertex to the object or ground. We have a small $w_d$ to make sure the gripper is close to the object or ground, and a large penetration weight $w_p$ to avoid collision. The optimized gripper pose is then given to the soft body simulation as initialization for data generation. 

\begin{wrapfigure}[24]{r}{0.45\textwidth}
  \vspace{-1.5em} 
  \begin{minipage}{\linewidth}
    \small
    \input{table/pseudo_code.tex}
  \end{minipage}
  \vspace{-2em}
\end{wrapfigure}

We perform a set of motions during the forward simulation: the tendons of the gripper first tighten with a fixed force, then the fingers move up with a fixed distance. We then record the ending object body force $f = [f_x, f_y, f_z, \tau_x, \tau_y, \tau_z] \in \mathcal{R}^6$, the object ending pose displacement with respect to the gripper $\Delta q = [\Delta q_x, \Delta q_y, \Delta q_z] \in \mathcal{R}^3$, and if there is ground collision $c_g \in \{0, 1\}$.

The neural network takes input of the partial point clouds, center of mass, and density of the object, and the 22D stiffness vector generated in the previous section. The partial point clouds are the object vertices within the bounding box of the gripper. The point clouds are fed into a 3-layer PointNet~\cite{qi2017pointnet} to extract features. The Pointnet is then connected with a 5-layer multilayer perceptron (MLP) to minimize the L1-loss of the ending body force, pose, and ground collision.

\subsubsection{Co-design of Stiffness and Pose}

We optimized the design and poses for each individual object, or jointly optimized for all the objects within the dataset. Denote the pose of the gripper in all the sampled candidate pose set $\mathcal{P}$ to be $\mathbf{p} \in \mathcal{P}$ and the stiffness vector to be $\mathbf{k}$.
The objective for an object set $\mathcal{O}$ is formulated as (all variables are conditioned on $\mathbf{p},\mathbf{k}$ but omitted for simplicity):
{\footnotesize
\begin{align*}
\mathcal{L}_{opt}(\mathbf p, \mathbf k) =
w_1\sum_{o \in \mathcal{O}} (\norm{f}+\norm{\Delta q})+ w_2\sum_{o \in \mathcal{O}} 
\bigl(\lvert\min(f_{y},0)\rvert
+ \lvert\min(\Delta q_{y},0)\rvert + c_{g}\bigr)
\end{align*}}
In Algorithm \ref{alg:joint_opt}, we begin with an initial set of candidate grasp poses and stiffness parameters, treating each object as its own group. At each iteration, we evaluate the loss for all pose–stiffness pairs within each group, sort them, and retain the top $B=5$ lowest losses (and their corresponding pose indices). We define the total loss as the sum of these $B$ losses across all objects, and perform a gradient‐descent update on the stiffness parameters via our smooth neural surrogate. Once the top‐$B$ losses for each group converge, we reduce each group to its single best pose and continue optimizing until all groups’ best losses stabilize. The algorithm then returns the optimized stiffness distribution and grasp pose for each object.

\section{Experiment Setup and Results}
We present a series of experiments both in simulation and on a real robotic platform. We evaluate our co-design optimized soft gripper in terms of: (a) efficiency of our proposed co-design framework; (b) generalization to objects with varying geometries and masses; and (c) comparative success rates across different object types against alternative gripper designs (rigid and super-soft). Additional experiments and results can be found in the Appendix.

\subsection{Simulation Experiments}
\input{table/sim_table.tex}
We first generated data to train the neural network surrogate. To ensure that the surrogate model accurately captured the underlying physics simulation, we sampled 50 random grasp poses for each object. For each pose, we further sampled 50 distinct stiffness values, resulting in approximately 80,000 valid data points (excluding samples where pose initialization fails). Data generation was executed on a server equipped with four 3090 GPUs and takes roughly one day to complete. 
We evaluated the success rate (SR) of grasping tasks on different objects. After grasping, a downward disturbance impulse of 0.1 N$\cdot$s was applied to the object. A  successful grasp was defined as the object being lifted and remaining securely held by the gripper: no ground collision, the object maintained contact with the gripper, and the object's body experienced non-zero force feedback. Success was determined over 850 frames recorded at 4000 fps.

\textbf{Jointly optimized gripper performs better.} We evaluated the optimized gripper (Appendix~\ref{sec:app_sim2real_transfer}) and tested whether the jointly optimized gripper outperforms all the other groups of grippers on both in-domain and out-of-domain objects, as shown in Table \ref{tab:simulation_success}. We tested four types of grippers: a soft gripper, a semi-rigid gripper, an individually optimized gripper based on each object, and a jointly optimized gripper across all in-domain objects, on both light (density = 2 kg/$m^3$) and heavy categories (density = 8 kg/$m^3$), which is chosen experimentally with stable simulation. 
In-domain data are 45 objects from the YCB~\cite{YCB} dataset, which are also used to generate data to train the neural network surrogate. Out-of-domain consists of 28 objects from the YCB (not in-domain), KIT~\cite{KIT}, and EGAD~\cite{EGAD} datasets. The out-of-domain objects differ significantly in size and shape from the in-domain ones. Evaluations are conducted under both fixed and sampled grasp poses. For the fixed-pose scenario, we select the first optimized candidate pose (corresponding to the best pose from Anygrasp). For the pose-sampling scenario, we select the best-performing candidate pose based on predictions from our neural network surrogate. All grippers achieve higher success rates when grasping lighter objects. Optimized grippers outperform the soft and semi-rigid grippers even on heavier objects. Furthermore, all grippers benefit from pose sampling, which samples the best pose from our neural network surrogate, with the jointly optimized gripper achieving the highest performance under this setting.

\begin{figure*}
\centering
\begin{subfigure}{0.245\textwidth}
\includegraphics[width=\textwidth]{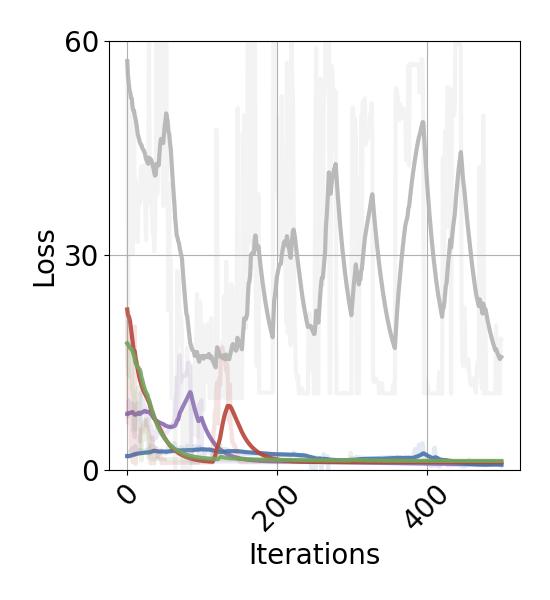}\vspace{-8pt}
\caption{}
\label{fig:diffsim_losses}
\vspace{-5pt}
\end{subfigure}
\begin{subfigure}{0.24\textwidth}
\includegraphics[width=\textwidth]{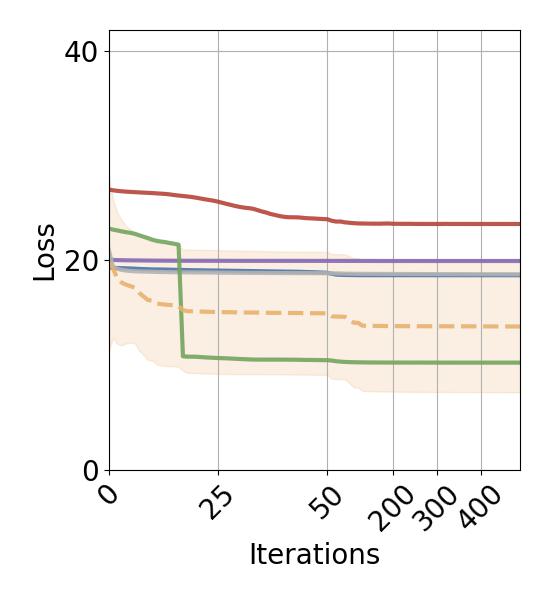}\vspace{-8pt}
\caption{}
\label{fig:opt_losses}\vspace{-5pt}
\end{subfigure}
\begin{subfigure}{0.24\textwidth}
\includegraphics[width=\textwidth]{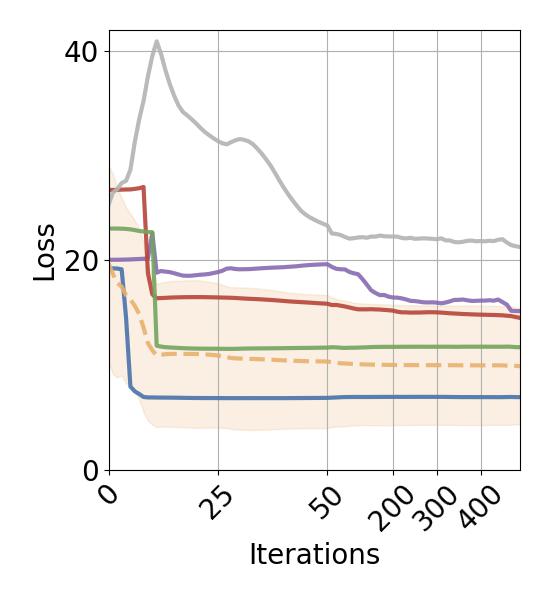}\vspace{-8pt}
\caption{}\label{fig:joint_losses}\vspace{-5pt}
\end{subfigure}
\begin{subfigure}{0.25\textwidth}
\includegraphics[width=\textwidth]{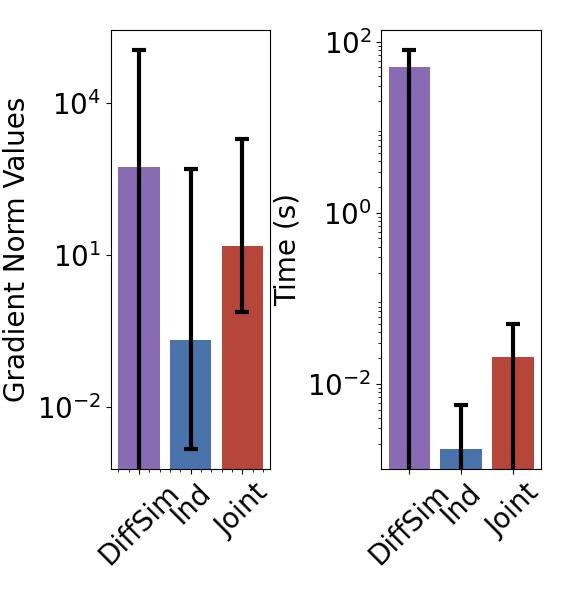}\vspace{-8pt}
\caption{}
\label{fig:bar}\vspace{-5pt}
\end{subfigure}
\caption{Numerical results of gradient-based optimization in simulations. (a) Differentiable simulation loss. (b) Individually optimized loss. (c) Jointly optimized loss. (d) The gradient norms and computation time of each iteration for differentiable simulation, individually optimized and jointly optimized.}
\label{fig:loss_compare}
\vspace{-8pt}
\end{figure*}

\textbf{Our co-design framework is efficient with smoother gradients compared with differentiable simulator gradients}.
During the optimization process, we conduct both individual and joint optimization, and compare them to the differentiable simulation-based optimization method. 
As shown in Figure \ref{fig:loss_compare}, compared with directly using gradients in the differentiable simulator, our neural physics-based optimization demonstrates significantly higher efficiency, yielding smoother gradients and smaller gradient norms, achieving a speedup of approximately three orders of magnitude in Figure~\ref{fig:bar}.
We visualize this by randomly selecting five objects and plotting their optimization losses in Figures \ref{fig:diffsim_losses} to \ref{fig:joint_losses}. The differentiable simulation losses were smoothed for clarity. The yellow dotted line represented the average loss across all in-domain objects, with the shaded region as the standard deviation.
A key observation is that joint optimization helped mitigate the risk of individual objectives getting trapped in local minima. Notably, several individual objects converged to lower loss values in joint optimization. By considering the losses of all objects simultaneously, the joint strategy had better convergence and resulted in a lower average loss across the dataset. We present our design optimized in simulation in the top right of Figure \ref{fig:architechture}, where the color from green to red represents the transition from hard to soft materials. An interesting finding is that most of the optimized grippers have at least one rigid tip, resembling the nails of primates.


\label{sec:result}

\subsection{3D Printing}
\begin{figure}
\centering
\begin{subfigure}{0.44\linewidth}
\includegraphics[width=\linewidth]{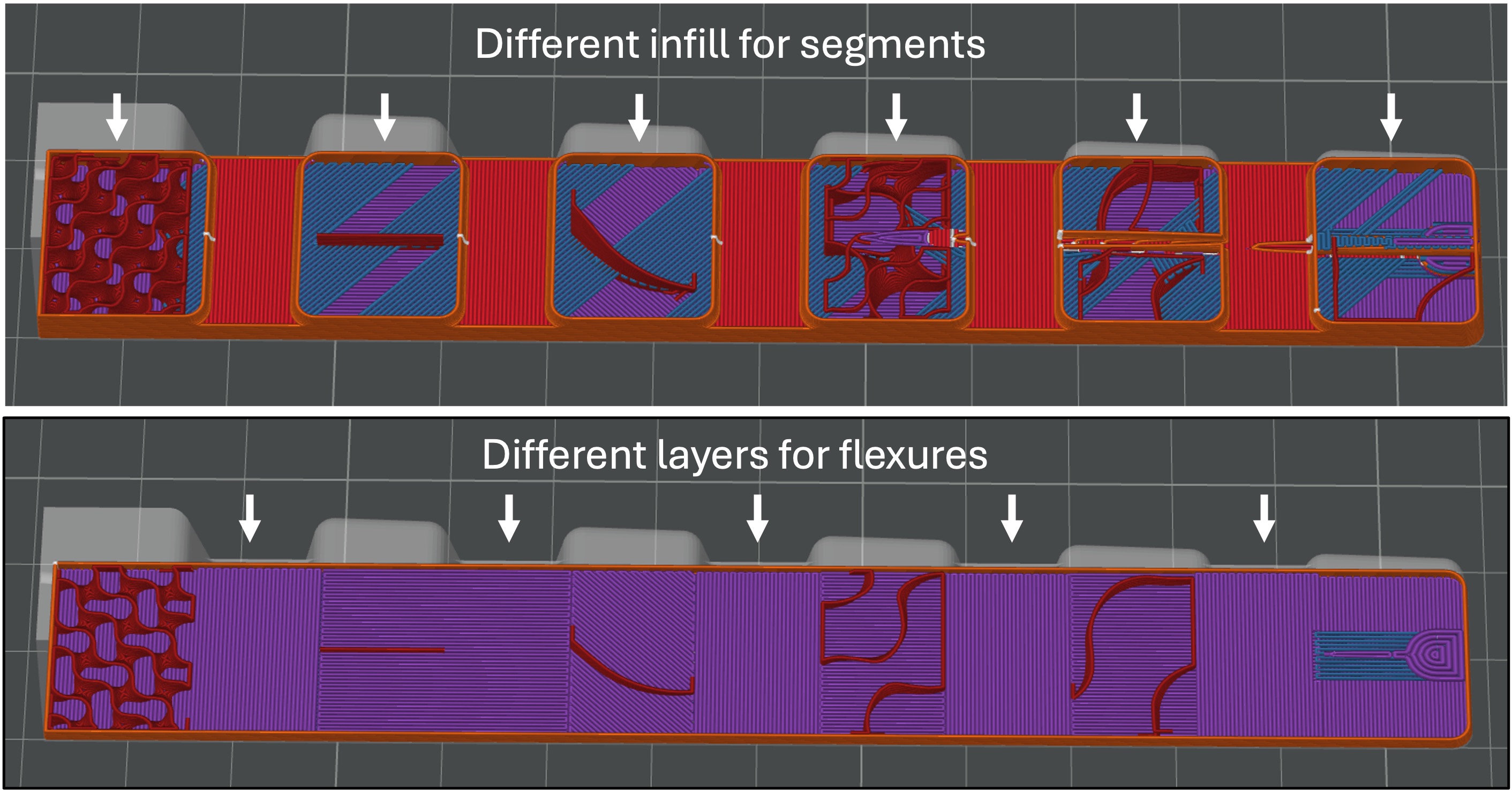}
\caption{}
\label{fig:print_infill}
\end{subfigure}
\begin{subfigure}{0.25\linewidth}
\includegraphics[width=\linewidth]{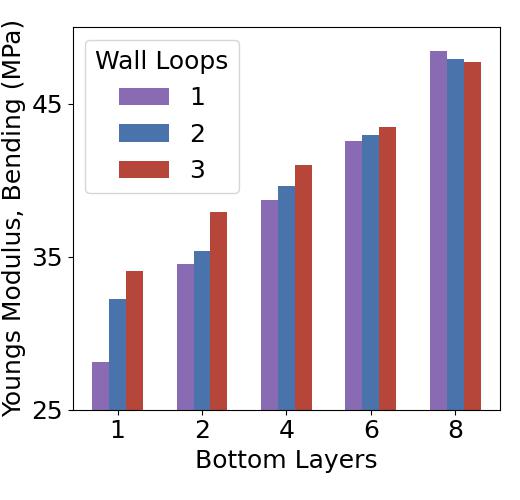}
\caption{}
\label{fig:k_flexure}
\end{subfigure}
\begin{subfigure}{0.265\linewidth}
\includegraphics[width=\linewidth]{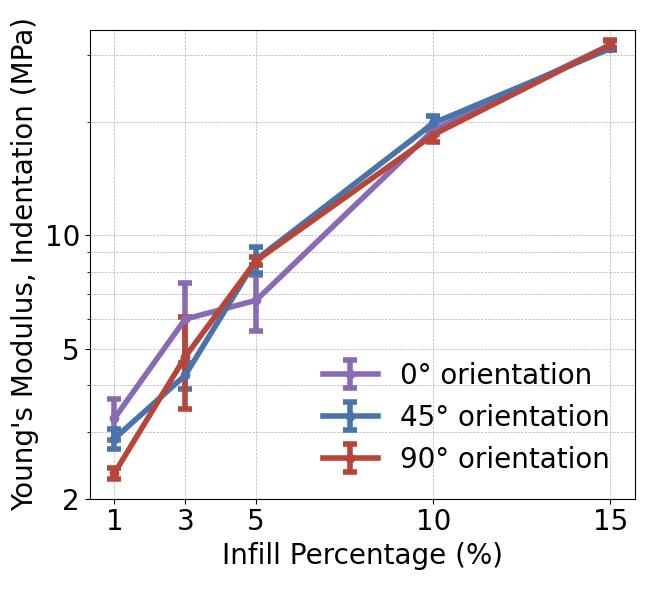}
\caption{}
\label{fig:k_finger}
\end{subfigure}
\caption{3D printing different stiffnesses. (a) Top: different infills for the segment blocks; Bottom: different numbers of layers for flexure blocks. (b) The Young's modulus of flexure blocks with a 3-point bending test. (c) Young's modulus of segment blocks from the indentation test.}
\label{fig:3d_print}
\vspace{-10pt}
\end{figure}

Different structural geometries and settings significantly influence the stiffness of soft bodies. As shown in Figure~\ref{fig:print_infill}, in 3D printing, the infill type and density, wall loops, top and bottom shell layers, printing speed, and temperature all matter. We transferred our simulation-optimized design to hardware by adjusting some of these settings in experiments. The flexure blocks were 2 mm thick with relatively small volume, thus, we modify the wall loops and shell layers. For segment blocks, we adjust the infill percentage and the infill printing directions. The deformation of the flexure based soft finger is predominantly depending on the relative stiffness. We see that for flexure blocks in Figure~\ref{fig:k_flexure}, one wall loop setup gives the largest range. For segment blocks in Figure~\ref{fig:k_finger}, the infill type combined with different infill directions gives a larger range. Therefore, we adapt these setups to print our simulation-optimized design.

\subsection{Real World Experiments}

Our hardware setup consisted of a 7-DoF Franka arm. We use two Intel RealSense D435 cameras: one mounted vertically on the end-effector (55 cm above workspace) and another positioned at a 30-degree inclination in the front-top workspace region (85 cm elevation). We mounted our design onto the Franka Hand, where a Dynamixel XL330 motor executed position-controlled tendon actuation. 

{\small
\input{table/real_table.tex}
}
We positioned the objects on a table and captured 1280 $\times$ 720 RGB and depth images using two Intel RealSense D435 cameras. These images were then input into Anygrasp to generate grasp pose candidates. Among these candidates, the optimal pose was selected using predictions from our neural network surrogate. We merged point clouds from both RealSense cameras, transformed them into the robot's base coordinate system, and extracted the partial point cloud of the target object using the bounding box defined by our gripper, which was later fed into our neural network surrogate. After determining the grasp pose, the gripper moved from its initial configuration to the target configuration, including the end effector's two prismatic joints connected to the gripper. We applied a consistent displacement of 1000 units (out of 4096 units per full revolution) to grasp. Finally, the robot lifted the gripper upward by 10 cm. A grasp attempt was considered successful if the robot lifted the object, securely held it without dropping it during translation.

We selected 10 objects with diverse geometries—including flat, cylindrical, box-shaped, irregular forms, and varying masses. A small black box was attached to the thin shovel to enhance its visibility for the depth camera. During our experiments, we observed distinct limitations associated with rigid and super-soft grippers when manipulating objects. Specifically, the SR of the super-soft gripper significantly decreased when handling heavier objects (mass greater than 100g), as it frequently failed to maintain a secure grasp despite achieving the target pose under tendon-driven control. Conversely, rigid grippers demonstrated enhanced performance with heavier objects, benefiting from their inherently stiffer structure. Our optimized gripper notably outperformed both rigid and super-soft grippers in handling heavier objects, particularly those presenting features amenable to deformation-based grasping. Moreover, for objects characterized by irregular geometries, such as watermelons, shovels, or bunny-shaped items, our optimized gripper and the soft gripper exhibited superior adaptability by effectively conforming to the object's contours, thereby achieving higher success rates. In contrast, the rigid gripper struggled to secure such irregularly shaped objects due to its limited contact surfaces, frequently resulting in grasp failures.

\section{Conclusion}
\label{sec:conclusion}
We have presented a uniform-pressure tendon routing model for flexure-based soft fingers, and a co‐design framework for soft grippers that jointly optimizes both the spatial distribution of stiffness and the grasping pose using a neural surrogate trained on high‐fidelity simulations. By replacing costly finite‐element evaluations with a fast, differentiable model, our method enabled efficient gradient‐based updates while retaining broad exploration via sample‐based initialization. Hardware experiments showed that the resulting gripper has better load‐bearing capacity and grasp stability compared to both rigid and highly compliant designs.
Looking forward, our surrogate‐driven pipeline can be extended to richer parameterizations and longer‐horizon manipulation tasks by integrating large action models to explore more diverse interactions. We anticipate that this combination of data‐driven optimization will generalize to a wide variety of soft robot systems, speed up the manufacturing cycle, and discover new, more generalizable designs.

\section{Limitations}\label{sec:limitations}

While our surrogate‐driven co‐design framework achieves fast, differentiable optimization over stiffness and grasp pose, it assumes a smooth, bounded design space and relies on an external pose sampler to first initialize candidate grasps for optimization. The diversity of the candidate grasping poses significantly influences the design. In our current sampling method, a significant portion of the poses are downward-facing. If we have a more diverse set of initializations that consists of horizontally positioned grasps, we can potentially explore directional stiffness in the designs. We currently restrict ourselves to varying material stiffness over a fixed geometry rather than co‐optimizing shape or topology, since unbounded or highly nonconvex geometry spaces is challenging in both learning accurate neural network surrogates and gradient‐based updates.

Many of the sampled grasping poses produced by our generator were kinematically infeasible in practice due to collisions or joint and workspace limits of the robot arm. In such cases, we simply discarded the failing pose and move to the next candidate, which can influence the network selector if no valid alternatives remain. Addressing this requires integrating collision‐aware sampling or on‐the‐fly pose refinement, and potentially repositioning the object within the robot’s reachable workspace.

The current training dataset is relatively limited. To obtain a reliable simulation of soft bodies interacting with rigid bodies, we required high quality dataset - the mesh is clean and water-tight. However, most of the available datasets we found did not match this requirement, or consisted of largely artificial and unrealistic objects. Training on larger, high-quality datasets can potentially improve the generalization of our approach.

%
\clearpage
\acknowledgments{This work was supported, in part, by NSF CCF-2112665 (TILOS), and gifts from Amazon and Meta. We thank the fruitful discussions with Eric Heiden, Nicklas Hansen, Xueyan Zou, Jiayao Yan, Yichen Zhai, Dongting Li, and Iman Adibnazari.}

\bibliography{example}  
\newpage
\section*{Appendix}
\subsection{Experiments with Disturbance}\label{sec:app_disturbance_exp}

We provide more justifications for larger impulses in simulation, shown in Table~\ref{tab:rebuttal_combined}(a). Here, 0.1 N.s is under the fixed disturbance force of 500 N for one frame. With larger impulses, the success rate decreases while the trend remains the same - the success rate is higher with pose sampling. However, with a significantly larger impulse, the sampled pose policy does not make a difference.
\input{table/sim_table_rebuttal}

\subsection{Role of Surrogate Model}\label{sec:app_surrogate_model}

We have optimized our simulation efficiency: generating 80k data takes around one day on six 3090 GPUs. Training the surrogate network on one 3090 GPU converges within 1 hour. We added experiments with reduced training data (uniformly sampled subset of 20k) and evaluated on light objects using a soft gripper. The 80k surrogate model converges to a loss of 0.1411, while the 20k surrogate converges to 0.1919. The success rate of this group decreases from 79.0\% to 69.6\%, indicating the crucial role of a surrogate model with high fidelity.

\subsection{Sim-to-real Transfer of Optimized Stiffness}\label{sec:app_sim2real_transfer}

In our simulation, the range of Young’s Modulus E is 0.7 MPa to 24 MPa. For numerical efficiency in simulation and optimization, we use the log E, which ranges from 13.5 to 17.0. Our soft gripper in simulation experiments has log E = 13.5, semi-rigid has log E = 16.5. Experimentally, softer or stiffer modulus beyond this range gives unstable simulation results. The optimized gripper values are shown in Table~\ref{tab:gripper_loge}. For hardware, we use the same material NinjaFlex Cheetah (TPU 95A) and modify only structural stiffness: infill for segments, bottom layers for flexures, as mentioned in the paper. Young’s modulus was measured with a Mark-10 force gauge mounted on a universal testing machine (UTM): three specimens per printer-parameter set were fabricated, and tested 3 times each, using three-point bending (flexure blocks) and indentation (segment blocks). The simulated stiffness values were mapped onto the printed fingers by linearly scaling them to the empirical modulus reported in Fig.5. Since all of the flexure blocks and segment blocks have the same geometry, respectively, the deformation behavior mainly depends on relative stiffness\cite{craig2020mechanics}. The optimized gripper showed that asymmetric stiffness and relatively rigid tips gave better grasp stability and robustness. We hope these insights could inspire future end-effector designs.
\input{table/opt_values}

\subsection{Sim vs. Real Performance}\label{sec:app_sim2real_performance}

We added two more objects that are soft and deformable in Table~\ref{tab:success_rate_real_rebuttal}. Even for object hard to grasp, the optimized gripper still outperforms than both rigid and soft ones.
We added experiments with the best designs found in dataset. For density of 8.0 ± 0.1$g/cm^3$, the success rate is 66.7\%, while with optimized design and policy, the success rate is 89.5\%. For light object with density of 2.0 ± 0.1$g/cm^3$, the success rate is 93.3\% similar to the optimized result. However, the max deviation in best dataset designs is 1.79 - meaning one gripper works the best on one object can be 6 times stiffer than as that for another. Instead, our framework effectively found the best design for all objects. 
For pose sampling, the sim and real approaches are the same during deployment. The difference is in data generation, optimization and deployment. During data generation, we augment the Anygrasp poses to obtain more diverse poses. During optimization, we sample all Anygrasp poses and optimize the design within ranked poses. During deployment on both sim and real, our model is used as a selector conditioned on a given design - either with the optimized design, or other baseline design parameters. We added comparison of two overlapping objects in sim and real, as shown in Table~\ref{tab:rebuttal_combined}(b).

\input{table/real_table_rebuttal}
\end{document}

%% file: table/pseudo_code.tex
\footnotesize
\begin{algorithm}[H]
\caption{JointOptimize($\mathcal P,\,\mathbf k_0$)}\label{alg:joint_opt}
\KwIn{Pose sets $\mathcal P$ for each object, initial stiffness $\mathbf k_0$}
\KwOut{Optimized $\mathbf k^*$, poses $p^*$}
\BlankLine
\For{$i\in\mathrm{Objects}$}{%
$\mathrm{converge}_i \!\gets\! \mathrm{false},\mathrm{bestPose}_i \!\gets\! \emptyset,\mathrm{patience}_i \!\gets\! 0, \mathrm{prevBest}_i\!\gets\! \emptyset$ 
}
\For{$t\leftarrow1$ \KwTo $N$}{
  $\ell_{\rm total}\gets 0$
\For{$i\in\mathrm{Objects}$}{
{
  $\ell[\mathbf p]\gets \mathcal L(\mathbf p,\mathbf k)$ for all $\mathbf p \in \mathcal P_i$ 
}
$\pi\gets\mathrm{argsort}(\ell)$\;
\If{$\mathrm{prevBest}_i\; \mathrm{is}\; \pi[1:B]$}{
$\mathrm{patience}_i \mathrel{+}= 1$
}\Else{
$\mathrm{patience}_i\gets 0$
}
\If{$\mathrm{patience}_i\ge T_p$}{
$\mathrm{converged}_i\gets\mathrm{true}$\;
$\mathrm{bestPose}_i\!\gets\! \pi(1)$
}
\If{$\mathrm{converge}_i$}{$\ell_{\rm total}\mathrel{+}=\ell[\pi(1)]; \mathrm{continue}$}
\Else{
  $\ell_{\rm total} \mathrel{+}= \sum_{j=1}^B \ell\bigl[\pi(j)\bigr]$\; $\mathrm{prevBest}_i\!\gets\! \pi[1:B]$}

}
{\tcp{gradient‐step on $\mathbf k$}
$\mathbf k \leftarrow \mathbf k - \alpha\,\nabla_{\mathbf k}\,\ell_{\rm total}$}
}
\Return{$\mathbf k,\{\mathrm{bestPose}\}$}\;
\end{algorithm}

%% file: table/sim_table.tex
\begin{table}[tbp]
  \centering
    \caption{Simulation success rates (Light/Heavy) for in‐ and out‐of‐domain objects. }
  \label{tab:simulation_success}
  {\scriptsize                
  \setlength{\tabcolsep}{4pt} 
  \begin{adjustbox}{width=0.7\linewidth}
    \begin{tabular}{@{} l c  cc  cc @{}}
      \toprule
      \multicolumn{2}{c}{\textbf{Method}} 
        & \multicolumn{2}{c}{\textbf{In Domain}} 
        & \multicolumn{2}{c}{\textbf{Out of Domain}} \\
      \cmidrule(r){1-2} \cmidrule(r){3-4} \cmidrule(l){5-6}
      \textbf{Group} & \textbf{Pose sampling} 
        & \textbf{Light} & \textbf{Heavy} 
        & \textbf{Light} & \textbf{Heavy} \\
      \midrule
      \multirow{2}{*}{Soft}      & \xmark      & 69.6\% & 56.5\% & 63.6\% & 51.5\% \\
                                    & \cmark       & 79.0\% & 65.8\% &72.7\% & 51.5\% \\
      \midrule
      \multirow{2}{*}{Semi‐Rigid}&  \xmark   & 52.2\% & 37.0\% & 39.4\% & 15.1\% \\
                                    & \cmark       & 60.5\% & 39.5\% & 33.3\% & 18.3\% \\
      \midrule
      \multirow{2}{*}{Individually Optimized}
                                    & \xmark  & 88.9\% & 75.6\% &   —    &   —    \\
                                    & \cmark       & 87.0\% & 80.4\% &   —    &   —    \\
      \midrule
      \multirow{2}{*}{Jointly Optimized}
                                    & \xmark  & 89.1\% & 78.3\% & \textbf{84.8}\% & \textbf{63.6}\% \\
                                    & \cmark       &\textbf{92.1}\% & \textbf{89.5}\% & 81.8\% & 57.6\% \\
      \bottomrule
    \end{tabular}
  \end{adjustbox}
  }
\vspace{-15pt}
\end{table}

%% file: table/real_table.tex
\begin{table}[tbp]
  \centering
   \setlength{\tabcolsep}{4pt} 
  \renewcommand{\arraystretch}{1.3}
  \caption{Real success rates for test objects of different mass and geometry}
  \begin{adjustbox}{width=0.8\linewidth}
  \begin{tabular}{l*{10}{c}}
 
    \toprule
    
    Type
    & 
    \begin{minipage}[b]{0.067\textwidth}\centering
      \includegraphics[width=\linewidth]{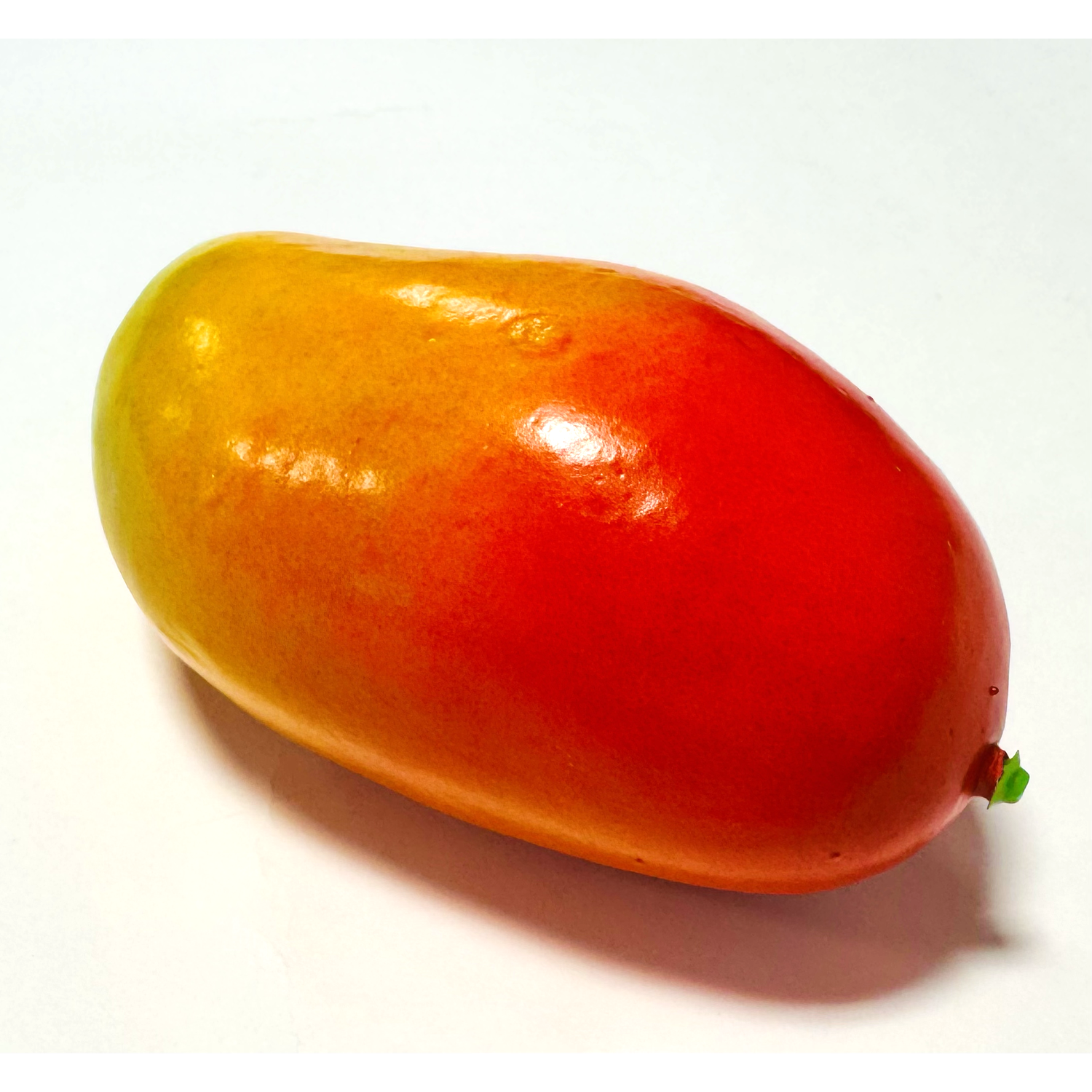}\\
      (15g)
    \end{minipage}
    & \begin{minipage}[b]{0.067\textwidth}\centering
      \includegraphics[width=\linewidth]{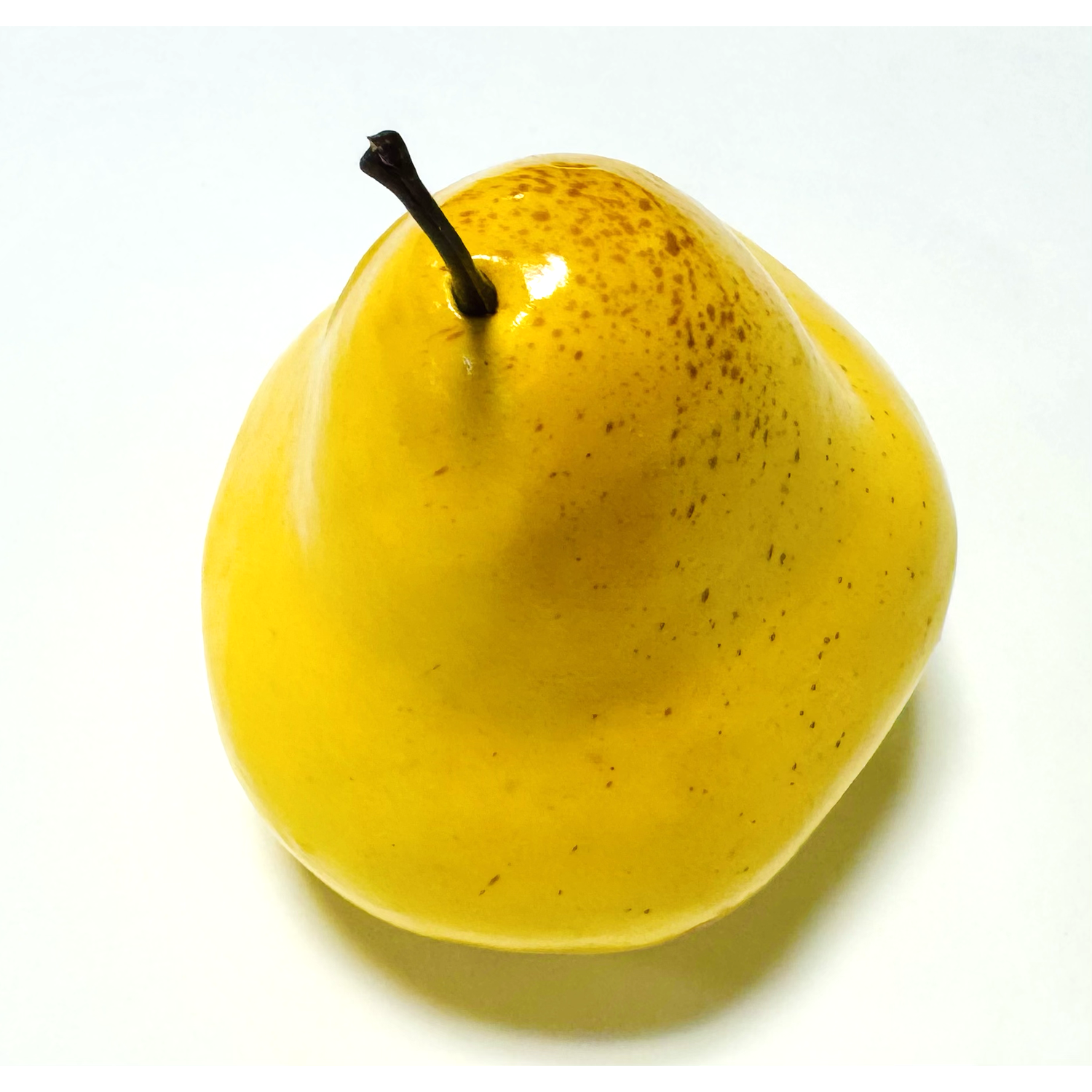}\\
    (15g)
    \end{minipage}
    & \begin{minipage}[b]{0.067\textwidth}\centering
      \includegraphics[width=\linewidth]{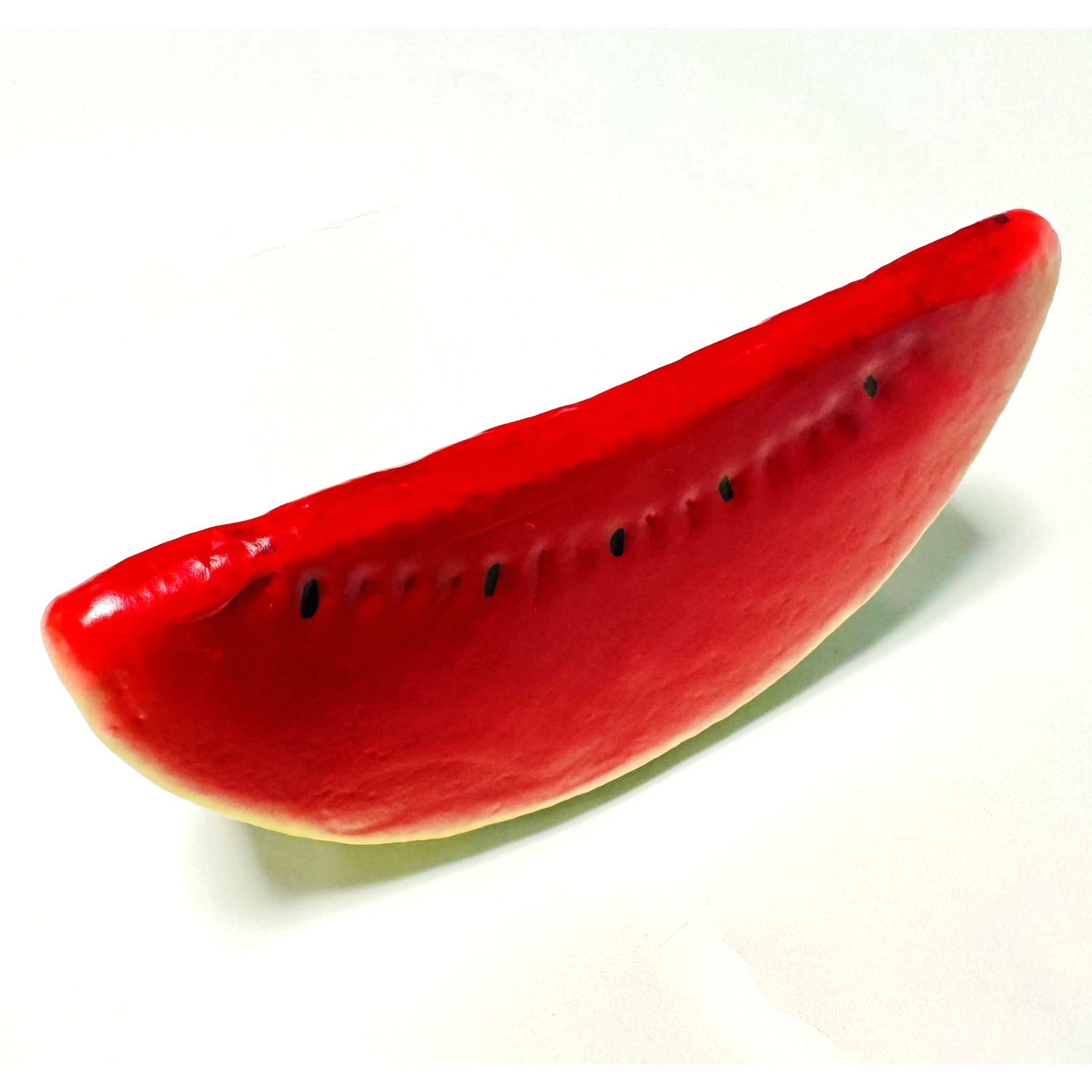}\\
     (15g)
    \end{minipage}
    & \begin{minipage}[b]{0.067\textwidth}\centering
      \includegraphics[width=\linewidth]{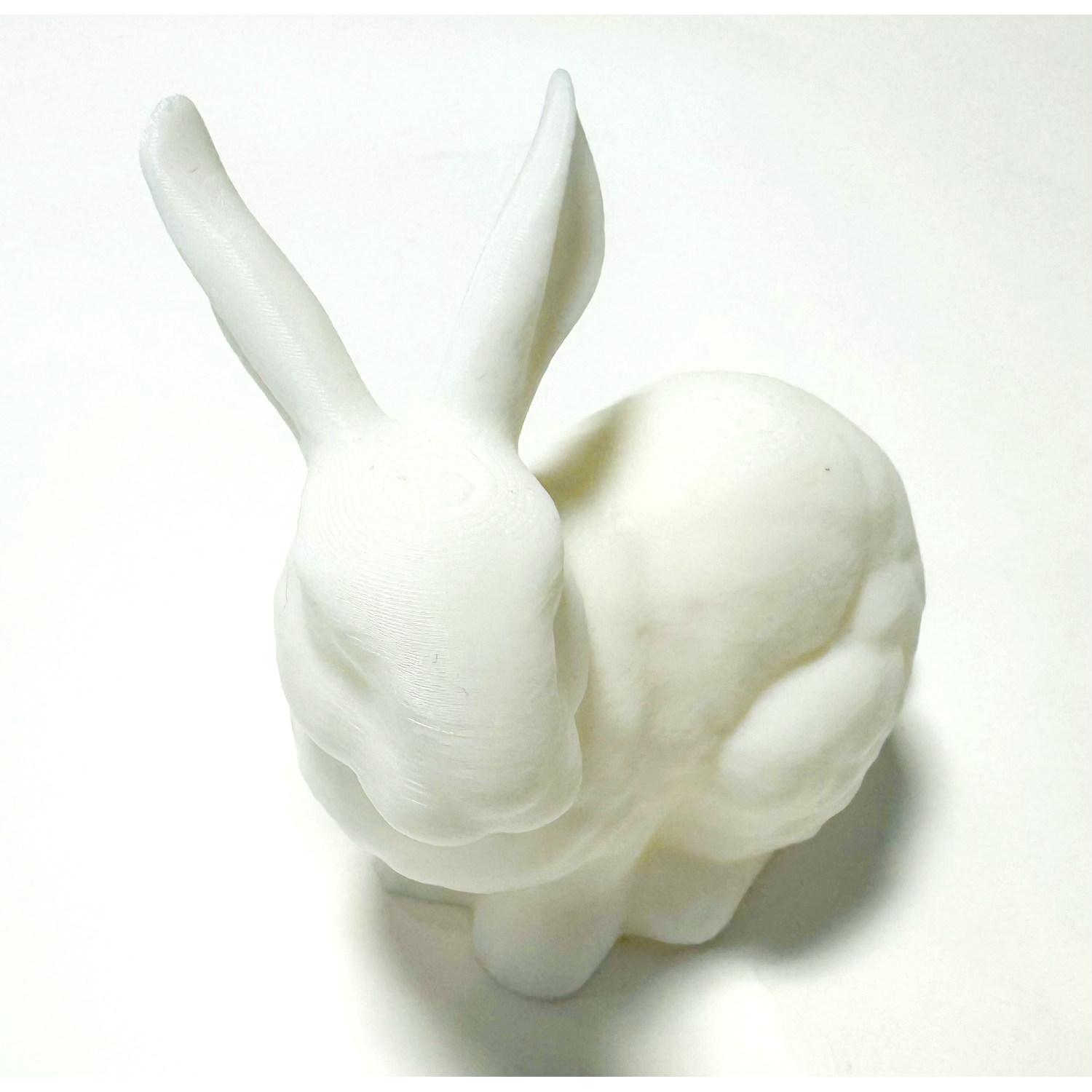}\\
     (37g)
    \end{minipage}
    & \begin{minipage}[b]{0.067\textwidth}\centering
      \includegraphics[width=\linewidth]{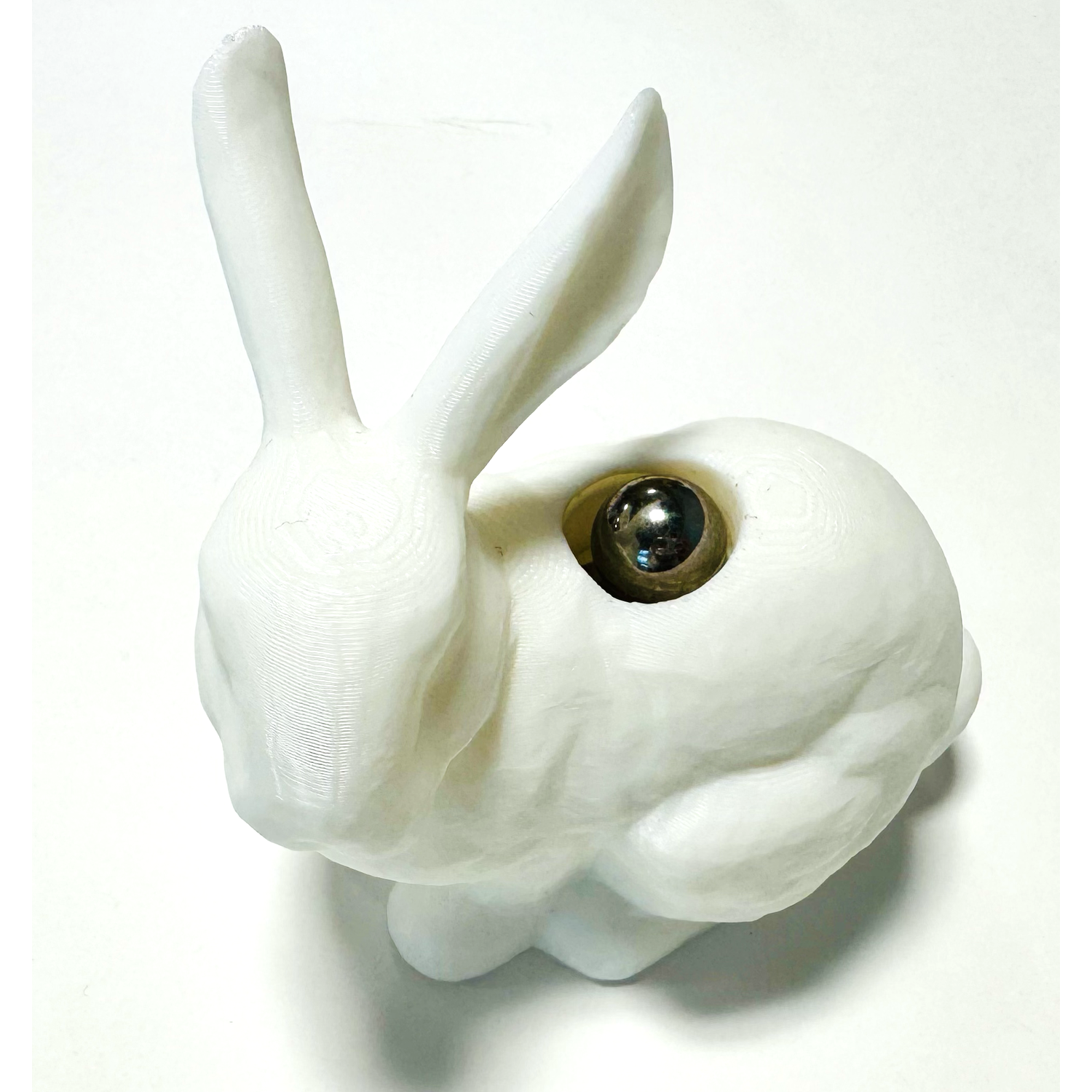}\\
     (101g)
    \end{minipage}
    & \begin{minipage}[b]{0.067\textwidth}\centering
      \includegraphics[width=\linewidth]{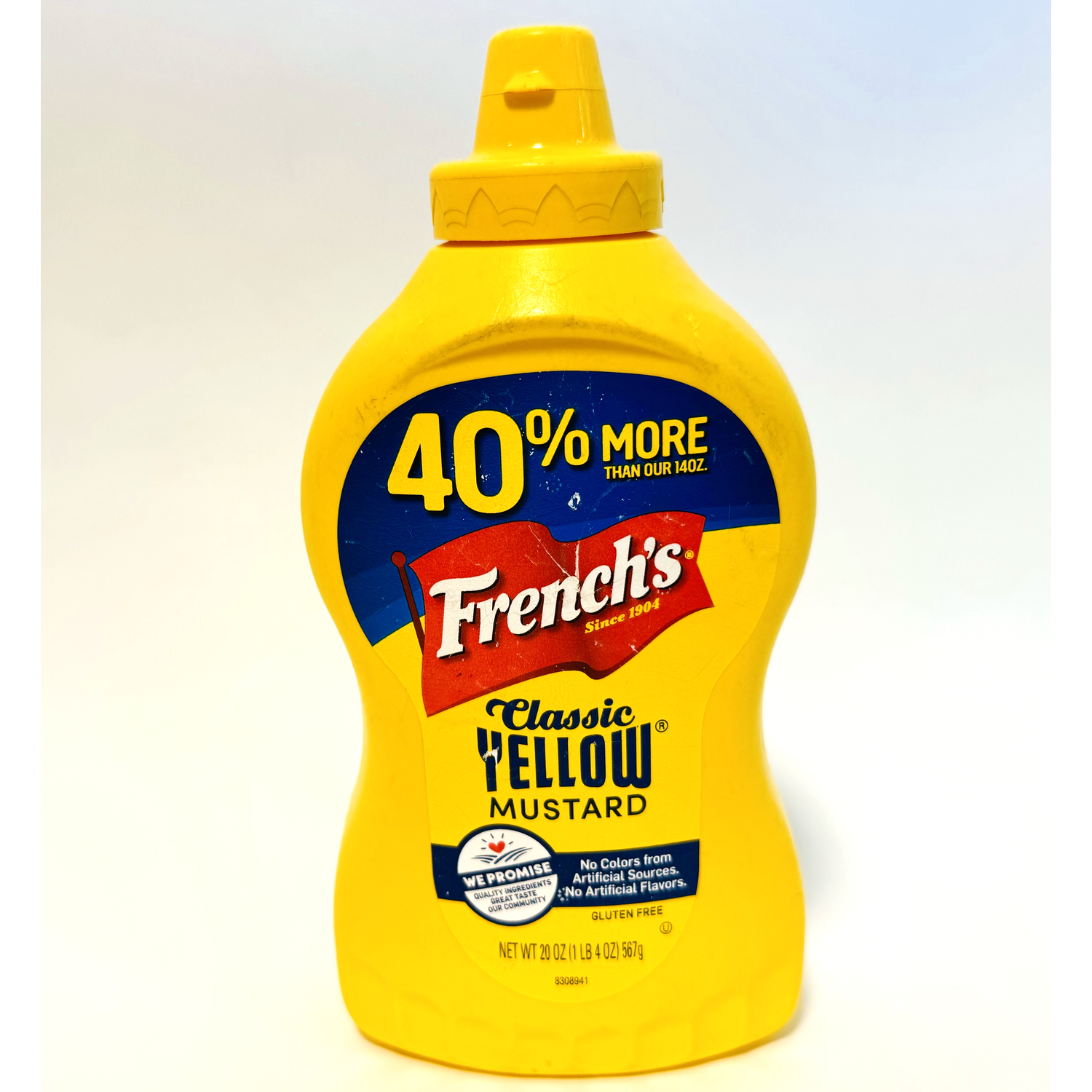}\\
       (47g)
    \end{minipage}
    & \begin{minipage}[b]{0.067\textwidth}\centering
      \includegraphics[width=\linewidth]{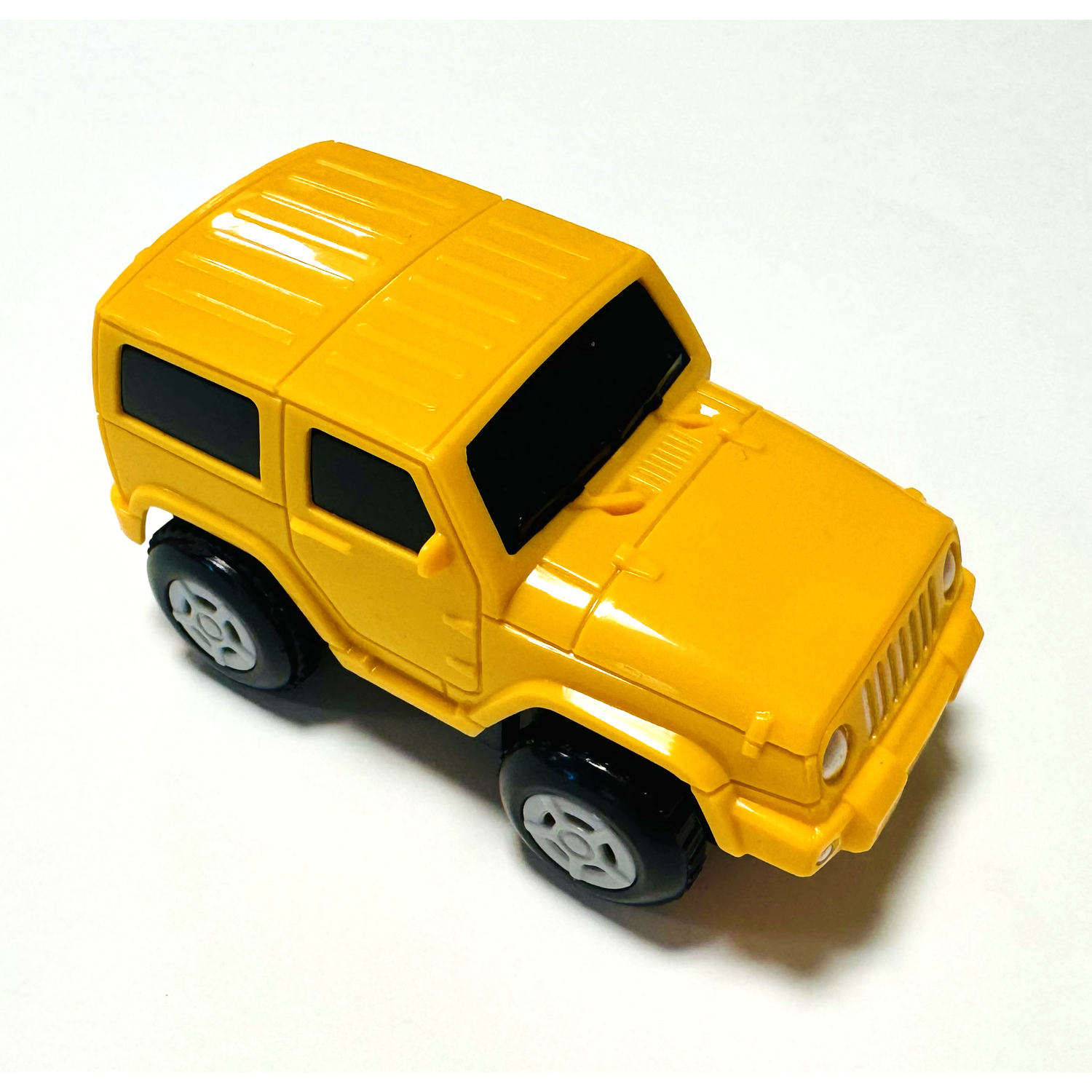}\\
       (52g)
    \end{minipage}
    & \begin{minipage}[b]{0.067\textwidth}\centering
      \includegraphics[width=\linewidth]{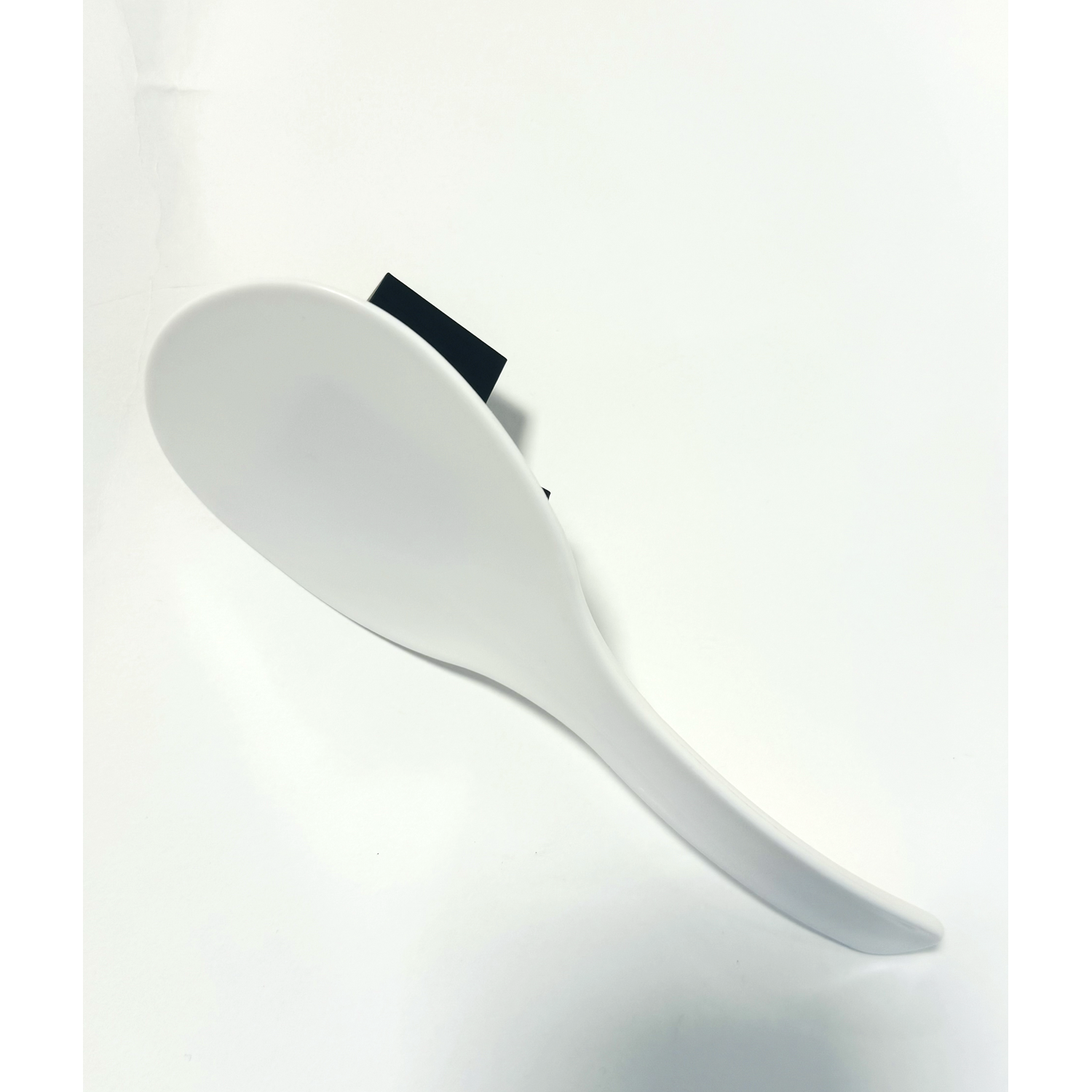}\\
      (41g)
    \end{minipage}
    
    & \begin{minipage}[b]{0.067\textwidth}\centering
      \includegraphics[width=\linewidth]{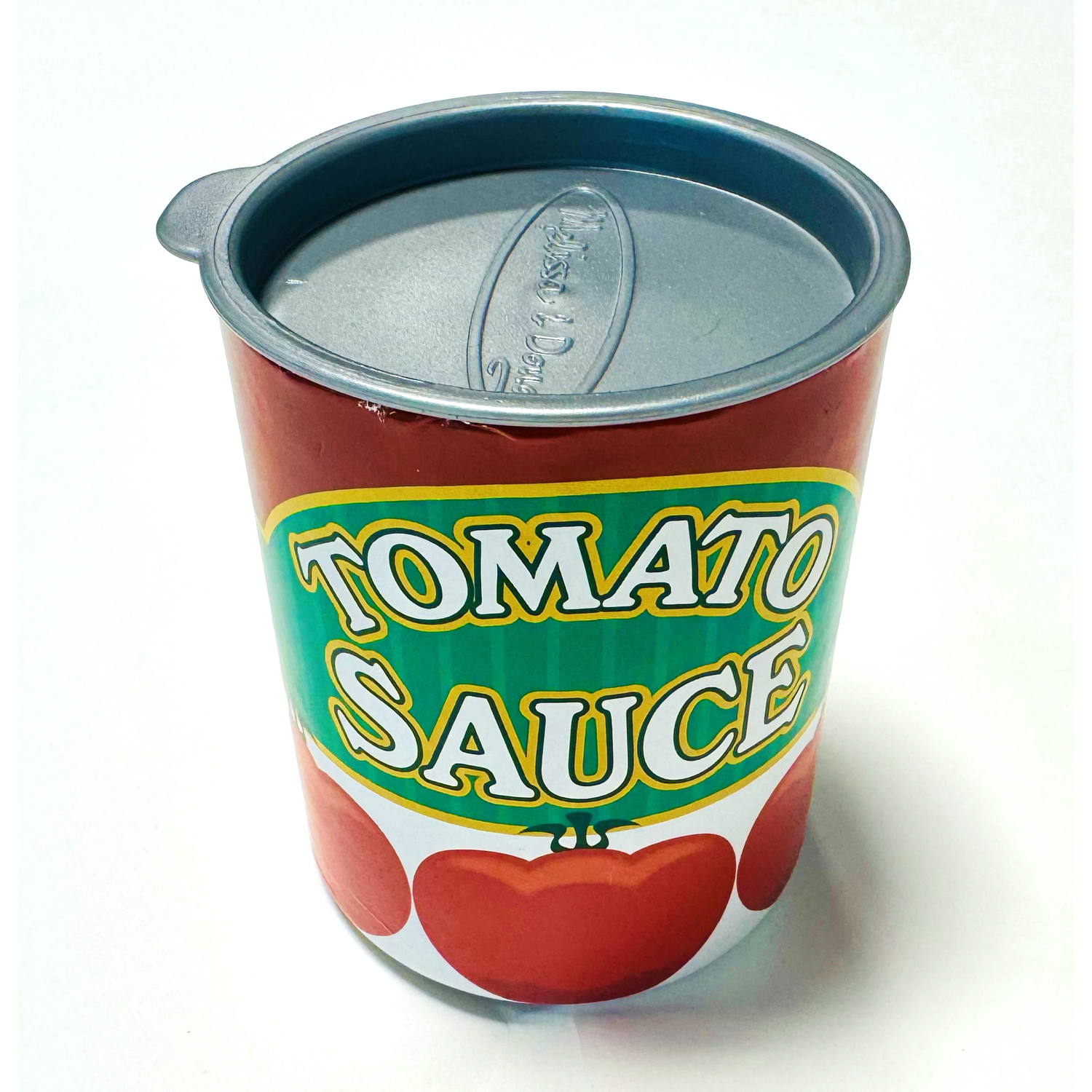}\\
       (121g)
    \end{minipage}
    & \begin{minipage}[b]{0.067\textwidth}\centering
      \includegraphics[width=\linewidth]{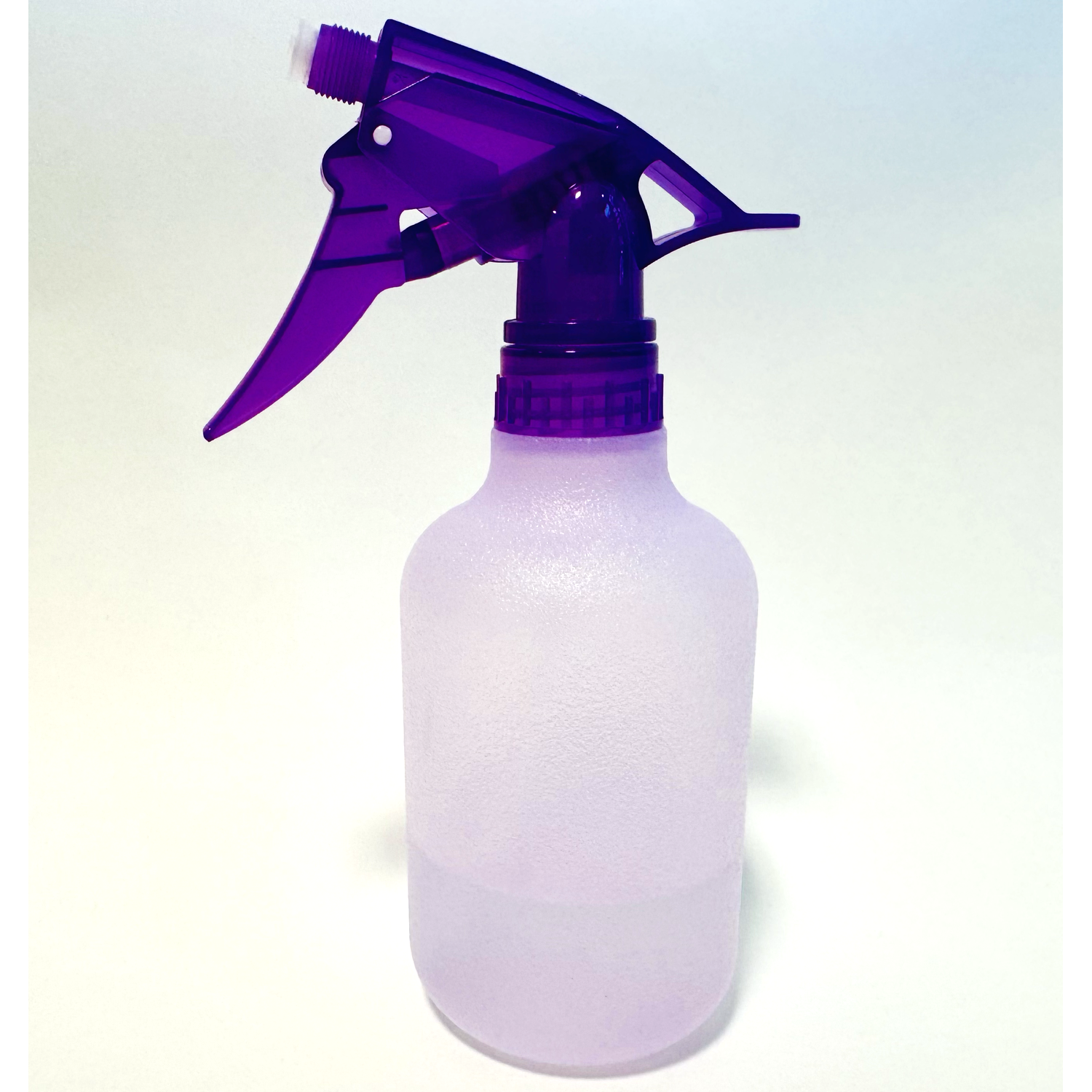}\\
       (180g)
    \end{minipage} \\
    \midrule
    Rigid     & \textbf{8/10} & \textbf{8/10} & 3/10 & 4/10 &3/10 &8/10 &6/10 &0/10 &5/10  & 5/10 \\
    Soft      & 7/10 & 7/10 & 5/10 &5/10 &2/10 &7/10 &1/10 &\textbf{6/10} &1/10 & 1/10 \\
    Optimized&6/10&7/10& \textbf{7/10} & \textbf{6/10} &\textbf{5/10} & \textbf{9/10}&\textbf{8/10}&5/10&\textbf{9/10}&\textbf{6/10}  \\
    \bottomrule
  \end{tabular}
  \end{adjustbox}
  \label{tab:success_rate_real}
  \vspace{-15pt}
\end{table}

%% file: table/sim_table_rebuttal.tex
\begin{table}[ht]
  \centering
  \vspace{-10pt}
  \caption{
  Success rates under disturbances and workspaces.}
  \label{tab:rebuttal_combined}
  \vspace{1pt}
  {\scriptsize
  \setlength{\tabcolsep}{4pt}
  \begin{minipage}[t]{0.48\linewidth}
    \centering
    \textbf{(a) Disturbance Robustness}\\
    \begin{adjustbox}{width=\linewidth}
      \begin{tabular}{@{}l c c c c@{}}
        \toprule
        \textbf{Group} & \makecell{\textbf{Pose}\\\textbf{sampling}} & \textbf{0.1 N$\cdot$s} & \textbf{0.2 N$\cdot$s} & \textbf{0.4 N$\cdot$s} \\
        \midrule
        \multirow{2}{*}{Soft} & \xmark & 69.6\% & 65.2\% & 67.4\% \\
                              & \cmark & 79.0\% & 76.1\% & 67.4\% \\
        \bottomrule
      \end{tabular}
    \end{adjustbox}
  \end{minipage}
  \hfill
  \begin{minipage}[t]{0.45\linewidth}
    \centering
    \textbf{(b) Sim vs. Real Performance}\\
    \begin{adjustbox}{width=\linewidth}
      \begin{tabular}{@{}l c c c c@{}}
        \toprule
        \textbf{Group} & \textbf{Workspace} & \textbf{Pear} & \makecell{\textbf{Mustard}\\\textbf{Bottle}} \\
        \midrule
        \multirow{2}{*}{\makecell{Jointly\\ Optimized}}  & Sim & 8/10 & 10/10 \\
                                           & Real & 7/10 & 9/10 \\
        \bottomrule
      \end{tabular}
    \end{adjustbox}
  \end{minipage}
  }
  \vspace{-5pt}
\end{table}

%% file: table/opt_values.tex
\begin{table}[ht]
  \centering
  \vspace{-10pt}
  \caption{
  Optimized gripper $\log(E)$ values. The values in parentheses $(\cdot)$ represent flexure values, others represent segment block values. The leftmost value is for the base block, and the rightmost is for the fingertip block.}
  \vspace{0.2em}
  \label{tab:gripper_loge}
  {\scriptsize
  \setlength{\tabcolsep}{4pt}
  \begin{adjustbox}{width=\linewidth}
    \begin{tabular}{@{}l*{10}{c}@{}}
      \toprule
      16.05 & (15.16) & 14.62 & (14.89) & 15.31 & (14.62) & 15.85 & (14.65) & 15.23 & (15.34) & 15.22 \\
      15.54 & (14.53) & 14.27 & (14.81) & 15.00 & (15.97) & 14.54 & (15.14) & 14.95 & (15.31) & 15.47 \\
      \bottomrule
    \end{tabular}
  \end{adjustbox}
  }
  \vspace{-5pt}
\end{table}

%% file: table/real_table_rebuttal.tex

\begin{table}[h]
  \centering
  \caption{ Additional objects success rates}
  \label{tab:success_rate_real_rebuttal}
  \setlength{\tabcolsep}{4pt}
  \begin{tabular}{@{} c c c c @{}}
    \toprule
    \textbf{Object / Mass} & \textbf{Rigid} & \textbf{Soft} & \textbf{Optimized} \\
    \midrule
    \raisebox{-0.5\height}{\includegraphics[width=0.08\linewidth]{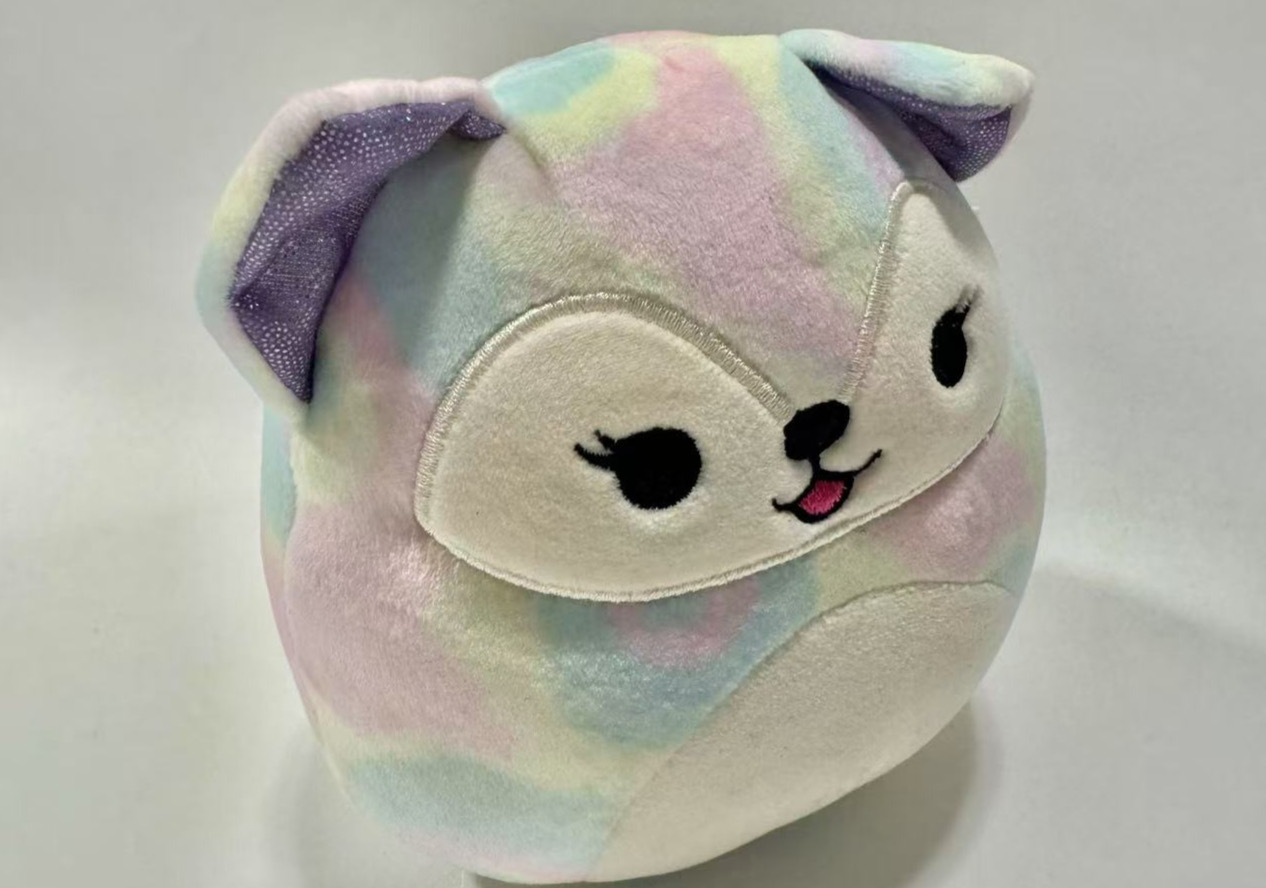}}~(57\,g)
      & 7/10 & 6/10 & \textbf{8/10} \\
    \raisebox{-0.5\height}{\includegraphics[width=0.08\linewidth]{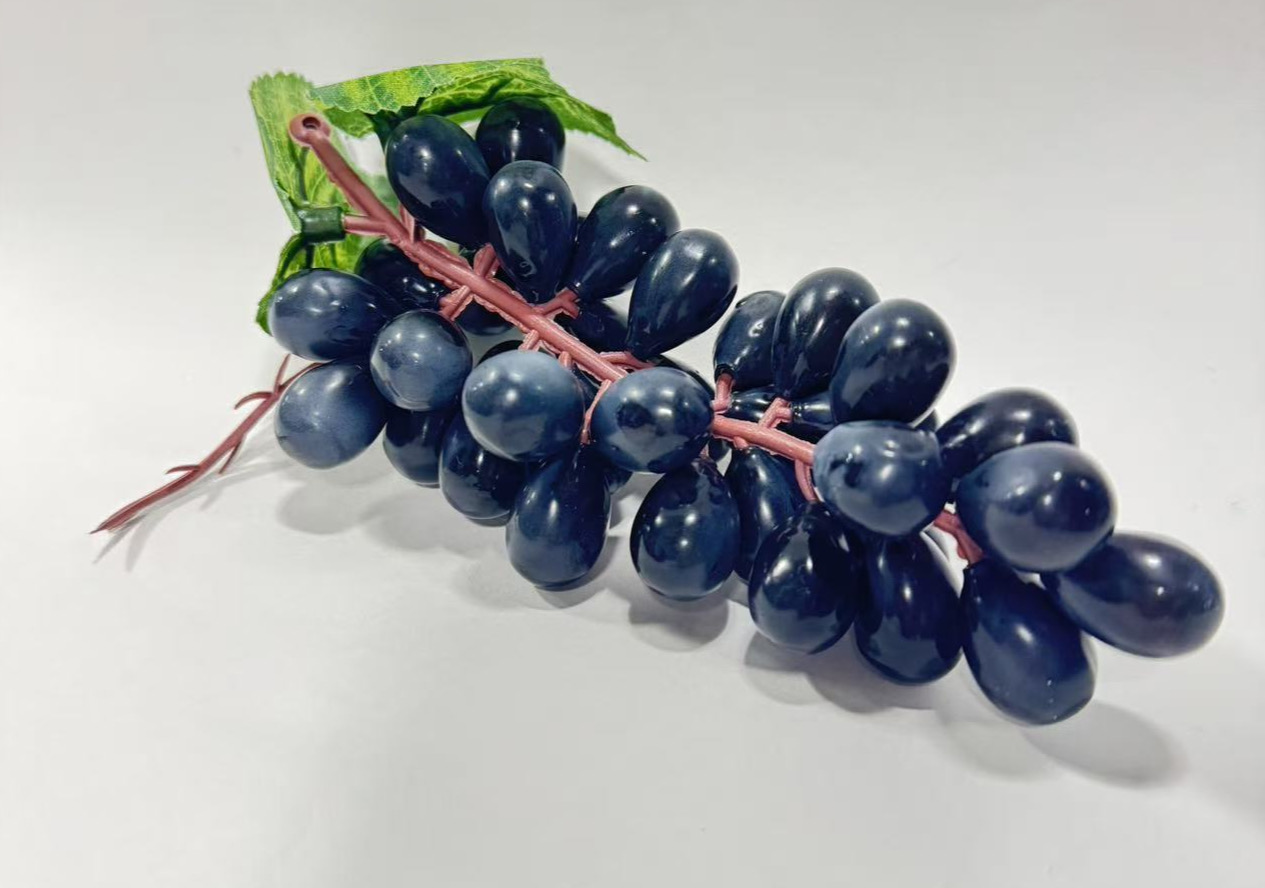}}~(69\,g)
      & 0/10 & 3/10 & \textbf{4/10} \\
    \bottomrule
  \end{tabular}
\end{table}